\documentclass{article}

\PassOptionsToPackage{numbers, compress}{natbib}

\usepackage[preprint]{neurips_2026}

\usepackage[utf8]{inputenc} 
\usepackage[T1]{fontenc}    
\usepackage{hyperref}       
\usepackage{url}            
\usepackage{booktabs}       
\usepackage{amsfonts}       
\usepackage{nicefrac}       
\usepackage{microtype}      
\usepackage{xcolor}         
\usepackage{amsmath}
\usepackage{amssymb}
\usepackage{algorithm}
\usepackage{algpseudocode}
\usepackage{wrapfig}
\usepackage{graphicx}
\title{Simple Extensions of Single-Objective Acquisition Functions and Hedge Strategies for Multi-Objective Bayesian Optimization}

\author{%
  Haris Moazam Sheikh\\
  Department of Aeronautical and Astronautical Engineering\\
  University of Southampton\\
  Southampton, United Kingdom \\
  \texttt{h.m.sheikh@soton.ac.uk} \\
}

\begin{document}

\maketitle

\begin{abstract}
Multi-objective Bayesian optimization (MOBO) is commonly approached through specialized acquisition functions or scalarization schemes designed to explicitly account for trade-offs among non-preferential objectives. In this work, we show that such complexity might be unnecessary. We propose a framework that extends standard single-objective acquisition functions directly to the multi-objective setting through a hypervolume-based methodology. We further extend hedge strategies for acquisition functions, which are typically used only in single-objective optimization, to the multi-objective regime. Our approach requires minimal modification to existing Bayesian optimization pipelines and avoids the need for bespoke multi-objective formulations. We demonstrate how a broad class of commonly used single-objective acquisition functions and hedge strategies can be adapted in a principled manner to handle multiple objectives, while preserving their intuitive interpretation and computational efficiency. Empirically, we evaluate the proposed methods across a range of synthetic and real-world multi-objective benchmarks. Despite their simplicity, our extensions consistently match or outperform more complex state-of-the-art MOBO methods in terms of optimization performance and sample efficiency. These results suggest that effective multi-objective Bayesian optimization can be achieved by reusing and carefully extending well-established single-objective acquisition strategies, offering a simpler and more flexible alternative to existing approaches.
\end{abstract}

\section{Introduction}

Modern scientific and engineering workflows increasingly require optimizing expensive black-box systems under tight evaluation budgets, where each query may correspond to a costly simulation, physical experiment, or deployment \cite{chen2019stiff,sadvours}. Bayesian optimization (BO) is well suited to this setting: it uses a probabilistic surrogate model and an acquisition function to select informative evaluations while balancing exploration and exploitation. Since the introduction of Efficient Global Optimization (EGO) \cite{Jones1998EGO}, BO has become a standard framework for sample-efficient optimization.

Many practical design problems, however, are inherently multi-objective. Engineers often care about competing criteria such as cost and reliability, or weight and safety, and preferences over these objectives are frequently unknown \emph{a priori}. In such non-preferential settings, there is generally no single best solution; the goal is instead to approximate the Pareto front of non-dominated trade-offs. Multi-objective Bayesian optimization (MOBO) therefore requires acquisition strategies that balance exploration and exploitation while promoting coverage across trade-offs in objective space.

Most existing MOBO methods address this challenge by designing acquisition functions that explicitly encode multi-objective structure, for example through scalarization \cite{Knowles2006ParEGO}, hypervolume improvement \cite{Daulton2020qEHVI,pmlr-v202-daulton23a}, or information-theoretic objectives \cite{HernandezLobato2016PESMO,Suzuki2020PFES,Tu2022JES}. These approaches are highly effective, but often require bespoke derivations, specialized approximations, or nontrivial implementation machinery. This work asks a simple question: do we need bespoke multi-objective acquisition machinery to obtain strong MOBO performance? We show that the answer can often be no. Rather than designing new multi-objective acquisitions, we propose a lightweight framework that reuses standard single-objective acquisitions and augments them with a Pareto-aware selection rule. Existing acquisitions generate promising candidates, while hypervolume-based ranking determines which evaluations best improve the current Pareto approximation. This yields a simple, modular approach that preserves the interpretability of classical BO acquisitions while enabling effective non-preferential MOBO.

A consistent observation in BO literature is that no single acquisition function performs best across all problem classes \cite{brochu2011portfolio}. Some acquisitions may excel on smooth objectives but struggle in highly multimodal landscapes; others may over-explore in sparse domains, or prematurely exploit suboptimal regions. As a result, acquisition choice often encodes implicit assumptions about the problem structure. In single-objective setting, this sensitivity has motivated hedge strategies that maintain a \textit{portfolio} of acquisition functions and adaptively select among them based on historical performance. This issue is arguably amplified in MOBO, where the method must additionally account for objective scaling, geometry, and Pareto-front coverage \cite{mixmobo}. 
Because our proposed extension enables diverse single-objective acquisitions with distinct exploration--exploitation behaviors to be instantiated within a unified MOBO pipeline at low computational cost, it naturally motivates portfolio-based selection. To address this, we introduce a multi-objective hedge strategy that adaptively selects among constituent MOBO policies using hypervolume-based historical performance, mitigating acquisition mismatch without committing to a single acquisition behavior \emph{a priori}.
Our contributions are summarized as:
\begin{enumerate}
\vspace{-0.185em}
    \item A simple hypervolume-ranked extension of standard single-objective acquisition functions to MOBO, based on Pareto candidate generation and posterior-mean hypervolume selection.
    \item A multi-objective hedge strategy that adaptively selects acquisition functions using hypervolume-based performance feedback.
    \item A comprehensive empirical evaluation on multi-objective benchmarks demonstrating that, despite their simplicity and minimal pipeline disruption, the proposed methods consistently match or outperform more complex state-of-the-art MOBO approaches.
\end{enumerate}

\section{Preliminaries}
\paragraph{Multi-Objective Optimization Problem:} We consider the problem of \textit{maximizing}\footnote{In this paper, unless stated otherwise, all objectives are assumed to be maximized.} a vector-valued objective function $\mathbf{f}:\mathcal{X}\to\mathbb{R}^M$, with $\mathbf{y}=\mathbf{f}(\mathbf{x})$, defined as
$\mathbf{f}(\mathbf{x}) = \big(f^{1}(\mathbf{x}), \dots, f^{M}(\mathbf{x})\big)$.
Here, $\mathbf{x} \in \mathcal{X} \subset \mathbb{R}^d$ is a $d$-dimensional decision vector, $\mathcal{X}$ is bounded, and the objectives $f^{m}(\mathbf{x}) \in \mathbb{R}$, $m \in \{1,\dots,M\}$, are assumed to be non-preferential. For non-preferential objectives, there generally does not exist a single \textit{optimal} solution. Instead, the goal is to identify the set of \textit{non-dominated} solutions, i.e., the Pareto front $\mathcal{P}^\star$ of objective vectors and the corresponding Pareto-optimal inputs $\mathcal{X}^\star$. An objective vector $\mathbf{f}(\mathbf{x})$ \textit{Pareto-dominates} another objective vector $\mathbf{f}(\mathbf{x}')$, denoted $\mathbf{f}(\mathbf{x}) \succ \mathbf{f}(\mathbf{x}')$, iff (i) $f^{m}(\mathbf{x}) \geq f^{m}(\mathbf{x}')$ for all $m\in\{1,\dots,M\}$, and (ii) there exists at least one $m\in\{1,\dots,M\}$ such that $f^{m}(\mathbf{x}) > f^{m}(\mathbf{x}')$. In this paper, $\mathrm{ND}(\cdot)$ is used to denote the non-dominated filtering operator in objective space. For any set $\mathcal{Y}\subset\mathbb{R}^M$,
$\mathrm{ND}(\mathcal{Y})=\{\mathbf{y}\in\mathcal{Y}\mid \nexists\,\mathbf{y}'\in\mathcal{Y}\text{ such that }\mathbf{y}'\succ\mathbf{y}\}$.

The \emph{Pareto front} and the corresponding set of \emph{Pareto-optimal inputs} are defined as
$\mathcal{P}^\star=\mathrm{ND}\big(\mathbf{f}(\mathcal{X})\big)$
and
$\mathcal{X}^\star=\{\mathbf{x}\in\mathcal{X}\mid\mathbf{f}(\mathbf{x})\in\mathcal{P}^\star\}$,
respectively, where $\mathbf{f}(\mathcal{X})=\{\mathbf{f}(\mathbf{x})\mid\mathbf{x}\in\mathcal{X}\}$. Over a continuous domain $\mathcal{X}$, $\mathcal{P}^\star$ is typically an infinite set. Given a finite set of evaluated points $\mathcal{S}\subset\mathcal{X}$, the approximate Pareto front is
$\mathcal{P}(\mathcal{S})=\mathrm{ND}\big(\mathbf{f}(\mathcal{S})\big)$,
where $\mathbf{f}(\mathcal{S})=\{\mathbf{f}(\mathbf{x})\mid\mathbf{x}\in\mathcal{S}\}$. The corresponding non-dominated subset of evaluated inputs is
$\mathcal{S}^\star=\{\mathbf{x}\in\mathcal{S}\mid\mathbf{f}(\mathbf{x})\in\mathcal{P}(\mathcal{S})\}$.

\paragraph{Multi-Objective Indicators:} To quantitatively assess the quality of a finite approximation $\mathcal{P}$ to the Pareto front, set-based performance indicators are usually employed, defined either in the objective space or the input space. Two widely used indicators are the \emph{hypervolume} ($HV$) and the \emph{inverted generational distance} ($IGD$). For a reference point $\mathbf{r}\in\mathbb{R}^M$ that is dominated by all objective vectors of interest, the hypervolume of $\mathcal{P}$ is the $M$-dimensional Lebesgue measure $\lambda(\cdot)$ of the region dominated by $\mathcal{P}$ and bounded by $\mathbf{r}$ from below:

\[
HV\left(\mathcal{P};\mathbf{r}\right)
=
\lambda\!\left(
\bigcup_{\mathbf{y}\in \mathcal{P}}
[\mathbf{r},\mathbf{y}]
\right),
\qquad
[\mathbf{r},\mathbf{y}]
=
\prod_{m=1}^M [r^{m}, y^{m}]
\]

IGD measures how well $\mathcal{P}$ \textit{covers} the Pareto front. Let $\mathcal{P}^\star_{\mathrm{disc}} \subset \mathcal{P}^\star$ denote a discrete representation of the true Pareto front; then IGD is defined as the average Euclidean distance from each $\mathbf{y}^\star\in\mathcal{P}^\star_{\mathrm{disc}}$ to its nearest neighbor $\mathbf{y} \in \mathcal{P}$:

\[
IGD\big(\mathcal{P};\mathcal{P}^\star_{\mathrm{disc}}\big)
=
\frac{1}{|\mathcal{P}^\star_{\mathrm{disc}}|}
\sum_{\mathbf{y}^\star\in\mathcal{P}^\star_{\mathrm{disc}}}
\min_{\mathbf{y}\in \mathcal{P}}
\left\lVert \mathbf{y}^\star-\mathbf{y} \right\rVert_2
\]

When $\mathcal{P}^\star$ is not known, $\mathcal{P}^\star_{\mathrm{disc}}$ is commonly replaced by a high-quality approximation (often the non-dominated set obtained from a global optimizer with a large evaluation budget).

\paragraph{Bayesian Optimization:} Bayesian optimization (BO) \cite{brochu2010tutorial} addresses the problem of optimizing an expensive black-box objective over a bounded domain $\mathcal{X}\subset\mathbb{R}^d$ by placing a Bayesian prior over the unknown function and selecting evaluation points sequentially. In the single-objective case, BO models $y=f(\mathbf{x}):\mathcal{X}\to\mathbb{R}$, typically with a Gaussian process (GP) prior $f \sim \mathcal{GP}(\mu(\cdot), k(\cdot,\cdot))$, where $k(\cdot,\cdot)$ is a symmetric positive semi-definite covariance kernel function. Given $n_t$ noisy observations at iteration $t$, $\mathcal{D}_t=\{(\mathbf{x}_i, y_i)\}_{i=1}^{n_t}$, with $y_i=f(\mathbf{x}_i)+\varepsilon_i$ (typically $\varepsilon_i\sim\mathcal{N}(0,\sigma^2)$), the posterior latent predictive distribution at any $\mathbf{x}\in\mathcal{X}$ is Gaussian, $p(f(\mathbf{x})=z \mid \mathcal{D}_t) = \mathcal{N}\big(z;\mu_t(\mathbf{x}), \sigma_t^2(\mathbf{x})\big)$, where $\mu_t(\cdot)$ and $\sigma_t^2(\cdot)$ are the GP posterior mean and variance, and $z \in \mathbb{R}$ is the latent variable. BO chooses the next query by maximizing an acquisition function $\alpha_t :\mathcal{X}\to\mathbb{R}$ (e.g., expected improvement, probability of improvement, or upper confidence bound) constructed from the posterior, i.e., $\mathbf{x}_{n_t+1}\in\arg\max_{\mathbf{x}\in\mathcal{X}} \alpha_t(\mathbf{x})$, then augments the dataset $\mathcal{D}_{t+1}=\mathcal{D}_t\cup\{(\mathbf{x}_{n_t+1}, y_{n_t+1})\}$ and updates the posterior; the exploration-exploitation trade-off is encoded by $\alpha_t$ through the dependence on both $\mu_t$ and $\sigma_t$. In the multi-objective setting, given $\mathcal{D}_t=\{(\mathbf{x}_i,\mathbf{y}_i)\}_{i=1}^{n_t}$, with $\mathbf{y}_i \in \mathbb{R}^M$, MOBO typically assigns a GP surrogate to each objective (independently or via a multi-output GP) \cite{balandat2020botorch}, yielding posteriors $p(f^{m}(\mathbf{x})=z^{m}\mid \mathcal{D}_t) = \mathcal{N}(z^{m};\mu_{t}^m(\mathbf{x}), (\sigma_{t}^m)^{2}(\mathbf{x}))$. Acquisition then targets Pareto front identification rather than a single scalar optimum, steering evaluations toward regions that refine the Pareto front while retaining uncertainty-driven exploration.

\section{Related Work}

A brief overview is presented of current state-of-the-art MOBO acquisition strategies that aim to approximate the Pareto front for a vector-valued black-box objective under a limited budget of expensive evaluations. A discussion of classical single-objective (SO) acquisition functions and hedge strategies used in this paper is presented in Appendix~\ref{app:SO}.

\paragraph{Scalarization-Based Methods:}
A classical strategy for MOBO is scalarization, in which the vector-valued objective is reduced to a scalar objective function that can be optimized using standard single-objective BO \cite{Knowles2006ParEGO, golovin2020randomhypervolumescalarizationsprovable, pmlr-v162-daulton22a}. For non-preferential objectives, this is usually done through randomized weight vectors. ParEGO \cite{Knowles2006ParEGO} is a canonical example, employing randomized augmented Tchebycheff scalarizations to transform the normalized objective vector:
\[
g(\mathbf{x}) =
\min_m \gamma^m \big(f^m(\mathbf{x})-z^m\big)
\;+\;
\rho \sum_{m=1}^M \gamma^m \big(f^m(\mathbf{x})-z^m\big)
\]
where $\mathbf{z}$ is a dominated reference point, $\boldsymbol{\gamma}$ is a randomly sampled weight vector, and $\rho > 0$ is a small augmentation parameter. In each BO iteration, a surrogate model is fitted to the scalar function $g$, which is then optimized using a standard SO acquisition function. Randomizing $\boldsymbol{\gamma}$ across iterations promotes exploration of different trade-offs along the Pareto front. Scalarization-based methods are attractive due to their simplicity and ease of implementation, but their performance depends strongly on choice of scalarization scheme and objective scaling. Collapsing objectives into a single scalar can also lead to uneven coverage of complex Pareto front geometries, particularly in higher dimensions.

\paragraph{Hypervolume-Based Methods:}
Another line of work extends BO to the multi-objective setting by leveraging Pareto quality indicators such as hypervolume to create acquisition functions \cite{Emmerich2006EHVI, Ponweiser2008}. An example is SMS-EGO \cite{Ponweiser2008}, which constructs \textit{potential} objective vectors $\mathbf{\hat{y}}_{pot}$ using upper-confidence bound, and then seeks to $\max_{\mathbf{x} \in \mathcal{X}} HV(\{\mathbf{\hat{y }}_{pot}\} \cup \mathcal{P}_{curr})-HV(\mathcal{P}_{curr})$. Here $\mathcal{P}_{curr}$ is the current Pareto front approximation. The candidate with the highest contribution is chosen for evaluation.

Hypervolume improvement-based acquisition functions optimize the expected gains in dominated hypervolume. One of the most widely used indicator-based acquisition strategies is the Expected Hypervolume Improvement (EHVI) \cite{Emmerich2006EHVI}, which defines the acquisition at iteration $t$ as
\[
\alpha_{\text{EHVI}}(\mathbf{x})
=
\mathbb{E}\!\left[
\Delta HV\big(\mathbf{f}(\mathbf{x}) \mid \mathcal{P}_t\big)
\;\middle|\;
\mathcal{D}_t
\right]
\]
In recent years, Daulton et al. have extended EHVI to accommodate parallel batch BO evaluations (\textit{q}EHVI)  \cite{Daulton2020qEHVI} and noisy black-box observations (\textit{q}NEHVI) \cite{DBLP:journals/corr/abs-2105-08195}. These methods enjoy attractive properties, including guarantees derived from the monotonicity and submodularity of hypervolume, but typically require nontrivial machinery such as box decompositions, Monte Carlo estimation, or integration over dominance regions.

More recently, Hypervolume Knowledge Gradient (HVKG) \cite{pmlr-v202-daulton23a} introduced a decision-theoretic look-ahead approach that selects evaluation points by maximizing the expected \emph{future} $(t+1)$ hypervolume under a one-step Bayes-optimal policy using
\[
\alpha_{\text{HVKG}}(\mathbf{x})
=\mathbb{E}_{\mathbf{y} \sim p(\mathbf{y} \mid \mathbf{x}, \mathcal{D}_t)}
\Big[\max_{\mathbf{x}'} HV\big(\mathcal{P}_{t+1}(\mathbf{y})\big)\Big]-HV(\mathcal{P}_t)
\]
Here, $\mathcal{P}_t$ is the current hypervolume-maximizing posterior-mean Pareto finite approximation. While HVKG is grounded in Bayesian decision theory, it entails nested expectations and inner optimizations over future Pareto structures, requiring carefully tuned sample-average approximation to remain tractable and one-shot optimization, which can be costly as the finite Pareto approximation grows.

\paragraph{Information-Theoretic Methods:}
A complementary class of MOBO methods formulates acquisition design as an information-gathering problem, selecting evaluations that maximize expected reductions in uncertainty, \textit{information gain} $I$, about Pareto-optimal structures. These methods generally take the form
\[
\alpha(\mathbf{x})
=
I\!\left( \mathbf{y}(\mathbf{x}) \,;\, \mathcal{Z^\star} \mid \mathcal{D}_t \right)
=
H(\mathcal{Z^\star} \mid \mathcal{D}_t)
-
\mathbb{E}_{\mathbf{y} \sim p(\mathbf{y} \mid \mathbf{x}, \mathcal{D}_t)}
\!\left[
H(\mathcal{Z^\star} \mid \mathcal{D}_t \cup \{(\mathbf{x},\mathbf{y})\})
\right],
\]
where $H$ is differential entropy and $\mathcal{Z}^\star$ denotes a latent Pareto-optimal object of interest (either Pareto-optimal inputs or Pareto-optimal objectives). Predictive Entropy Search for Multi-Objective Optimization (PESMO) \cite{HernandezLobato2016PESMO} targets information gain about the Pareto set \(\mathcal{X}^\star\), i.e., the set of Pareto-optimal inputs. Pareto Front Entropy Search (PFES) \cite{Suzuki2020PFES} instead focuses on the Pareto front \(\mathbf{f}(\mathcal{X}^\star)\) in objective space. Joint Entropy Search (JES) \cite{Tu2022JES} unifies these by maximizing information gain about both Pareto-optimal inputs and their corresponding objective values. While theoretically well motivated, information-theoretic approaches typically rely on sophisticated approximations such as truncated Gaussian sampling or Monte-Carlo integration over random Pareto sets to address the intractability of exact entropy computations, and the acquisition performance can depend on specific approximation choices and numerical settings.

\paragraph{MO Hedge Strategies:}
Hedge strategies for multi-objective Bayesian optimization have received limited attention. To our knowledge, HedgeMO is the only existing Hedge strategy specifically formulated for multi-objective optimization \cite{mixmobo}. The strategy maintains a portfolio of acquisition functions, each of which proposes a batch of candidate points. The historical nominees of each acquisition function are evaluated using the current surrogate posterior mean, and the resulting rewards are normalized separately for each objective and summed across objectives to define the Hedge reward for each strategy in the portfolio. The corresponding selection probabilities are then defined using the exponential weighting rule of single-objective GP--Hedge \cite{brochu2011portfolio}, 
described in Appendix~\ref{app:SO}. However, the additive reward collapses the objective-wise performance into a single scalar and therefore does not explicitly account for the Pareto quality of the historical nominees. 

We note that other single-objective acquisition-ensemble methods, such as MACE \cite{pmlr-v80-lyu18a} and DMEA \cite{chen2022dynamic}, could potentially be adapted to MOBO, provided that each constituent MOBO acquisition strategy returns a scalar utility that can be treated as an objective of the ensemble. Such an extension would, however, become increasingly difficult for large portfolios because every additional strategy introduces another objective into the auxiliary multi-objective optimization problem.



\section{Methodology}
\label{methods}
\subsection{Hypervolume‑Ranked Acquisition Extension}

We propose a simple approach, \textit{q}HAX-$\alpha$ (Parallel Hypervolume-ranked Acquisition Extension), that extends standard single-objective acquisition functions to the multi-objective setting via Pareto-diverse candidate generation and hypervolume-ranked selection. Importantly, \textit{q}HAX-$\alpha$ does not introduce bespoke multi-objective acquisition machinery; instead, it reuses existing single-objective acquisition functions and injects Pareto awareness at the decision stage, as summarized in
Algorithm~\ref{alg:qhax}.

\paragraph{Vector-Valued Acquisition:}
Consider a multi-objective black-box function $\mathbf{f}(\mathbf{x}) = \big(f^1(\mathbf{x}), \dots, f^M(\mathbf{x})\big)$, where all objectives are to be maximized. At each BO iteration $t$, we maintain an evaluation dataset $\mathcal{D}_t = \{(\mathbf{x}_i, \mathbf{y}_i)\}_{i=1}^{n_t}$, where $\mathbf{y}_i$ denotes noisy observations of $\mathbf{f}(\mathbf{x}_i)$. Independent Gaussian process models are fitted for each of the $M$ objectives, yielding posteriors $p(f^{m}(\mathbf{x}) = z^{m} \mid \mathcal{D}_t)$. In practice, the observations are standardized before model fitting.

Let $\alpha(\cdot)$ denote a standard single-objective acquisition function (e.g., Expected Improvement (EI), Probability of Improvement (PI), or Upper-Confidence Bound (UCB)). For each objective $m \in \{1,\dots,M\}$, at iteration $t$ we define 
\(\
\alpha^m_{t}(\mathbf{x})=\alpha\!\left(\mathbf{x};\, p(f^{m}(\mathbf{x}) = z^{m} \mid \mathcal{D}_t)\right),
\)
yielding a vector-valued acquisition
$
\boldsymbol{\alpha}_t(\mathbf{x})=\big(\alpha^1_{t}(\mathbf{x}), \dots, \alpha^M_t(\mathbf{x})\big)\in \mathbb{R}^M.
$

This construction preserves the semantics and exploration--exploitation behavior of the underlying single-objective acquisition for each objective. To explore trade-offs across objectives, an approximate Pareto-optimal candidate set is obtained by approximately solving
$
\max_{\mathbf{x} \in \mathcal{X}} \; \boldsymbol{\alpha}_t(\mathbf{x})
$
in the Pareto sense. In practice, any Pareto-front generating optimizer (e.g., NSGA-II~\cite{996017}) can be used to obtain a nondominated candidate set
$
\mathcal{C}_t = \{\mathbf{c}_{t,1}, \dots, \mathbf{c}_{t,J}\}.
$
This set spans trade-offs between different objectives. Unlike other strategies, \textit{q}HAX-$\alpha$ partially decouples the exploration--exploitation trade-off for each objective from the trade-off across objectives.

\paragraph{Candidate Selection through Hypervolume Ranking:}
Given the candidate set $\mathcal{C}_t$, the goal is to select the next evaluation point. 
To evaluate multi-objective utility, we construct a denoised Pareto front using the posterior-mean objective vector $\boldsymbol{\mu}_t(\mathbf{x}) = \big(\mu_t^1(\mathbf{x}),\dots,\mu_t^M(\mathbf{x})\big)$, where $\mu_t^m(\mathbf{x}) = \mathbb{E}[f^m(\mathbf{x})\mid\mathcal{D}_t]$.
The current Pareto front approximation is defined as
$
\mathcal{P}_t = \mathrm{ND}\big(\{\boldsymbol{\mu}_{t}(\mathbf{x}_i)\}_{i=1}^{n_t}\big).
$
For a candidate $\mathbf{c}_t \in \mathcal{C}_t$, we define its utility as
\[
h_t(\mathbf{c}_t)=HV\big(\mathcal{P}_t \cup \{\boldsymbol{\mu}_{t}(\mathbf{c}_t)\}; \mathbf{r}_t\big),
\]
where $HV(\cdot; \mathbf{r}_t)$ denotes the hypervolume indicator with respect to a reference point $\mathbf{r}_t$. 
The next evaluation point is selected as
$
\mathbf{c}_{t+1}=\arg\max_{\mathbf{c}_t \in \mathcal{C}_t} h_t(\mathbf{c}_t).
$
Since $h_t$ is used solely to rank candidates, $\mathbf{r}_t$ can be any point that is strongly dominated by all elements in $\mathcal{P}_t \cup \boldsymbol{\mu}_t(\mathcal{C}_t)$. If multiple candidates yield equal hypervolume values, the candidate ranked higher by the Pareto-front optimizer is selected, as such algorithms typically use crowding-distance-based tie-breaking.

\paragraph{Relation to Hypervolume-Based Acquisitions:}
Although $h_t(\mathbf{c}_t)$ is evaluated on deterministic posterior means $\boldsymbol{\mu}_t(\mathbf{c}_t)$, the selection is not purely exploitative because maximization is restricted to $\mathcal{C}_t$, which acts as an uncertainty-aware filter defined by the Pareto front of $\boldsymbol{\alpha}_t(\mathbf{x})$. This two-stage decomposition fundamentally distinguishes \textit{q}HAX-$\alpha$ from existing hypervolume acquisitions. Unlike EHVI-based strategies~\cite{Daulton2020qEHVI, Emmerich2006EHVI, DBLP:journals/corr/abs-2105-08195}, which couple exploration and trade-off selection by integrating hypervolume improvement over the predictive posterior (or latent frontier realizations), and unlike HVKG \cite{pmlr-v202-daulton23a}, which requires nested look-ahead expectations, \textit{q}HAX-$\alpha$ inherits exploration from $\boldsymbol{\alpha}_t(\mathbf{x})$ and evaluates hypervolume only on posterior means over a finite candidate set. Similarly, whereas SMS-EGO \cite{Ponweiser2008} substitutes optimistic UCB estimates $\mathbf{\hat{y}}_{pot}(\mathbf{x})$ directly into the $HV$-indicator across $\mathcal{X}$, coupling predictive variance with objective-space dominance geometry, \textit{q}HAX-$UCB$ confines optimism to generating $\mathcal{C}_t$ and ranks candidates using denoised posterior means $\boldsymbol{\mu}_t(\mathbf{c}_t)$ relative to $\mathcal{P}_t$.

\paragraph{\textit{q}-Batch Selection for Parallel Evaluations:}
\textit{q}HAX-$\alpha$ can be naturally extended to parallel batch selection by greedily re-ranking candidates after each selection. 
For a batch size $Q$, the top candidate $\mathbf{c}^{q+1}_{t+1}$ is selected at each inner iteration based on hypervolume, and the current Pareto approximation is updated as
$
\mathcal{P}_t^{q+1} \leftarrow \text{ND}(\mathcal{P}_t^q \cup \{\boldsymbol{\mu}_{t}(\mathbf{c}^{q+1}_{t+1})\}).
$
The selected candidate is then removed from $\mathcal{C}_t$, and the remaining candidates are re-ranked. This strategy is computationally efficient, as surrogate models are not refitted during batch construction, enabling \textit{q}HAX-$\alpha$ to scale to larger batch sizes. If $|\mathcal{C}_t| < Q$, which occurs rarely in practice, the remaining inputs can be sampled uniformly at random.

\paragraph{Theoretical Remark -- Submodularity-Based Guarantee:}
While other set functions (e.g., IGD) may be used heuristically for candidate ranking, hypervolume admits a simple theoretical justification for batch construction. For $Q=1$, maximizing $h_t(\mathbf{c}_t)$ over the finite set $\mathcal{C}_t$ is exact. For $Q > 1$, holding the posterior mean $\boldsymbol{\mu}_t$ fixed across the inner loop (rather than updating the surrogate via fantasization) ensures that the posterior-mean hypervolume gain $\Delta HV(\mathcal{B} \mid \mathcal{P}_t^0; \mathbf{r}_t) = HV\big(\mathcal{P}_t^0 \cup \boldsymbol{\mu}_t(\mathcal{B}); \mathbf{r}_t\big) - HV(\mathcal{P}_t^0; \mathbf{r}_t)$ is a stationary, monotone submodular set function over subsets $\mathcal{B} \subseteq \mathcal{C}_t^0$. Consequently, greedy sequential selection achieves a $(1 - 1/e)$-approximation to the optimal $Q$-candidate batch attainable within $\mathcal{C}_t^0$; equivalently, the instantaneous optimality gap relative to the best $Q$-point subset in $\mathcal{C}_t^0$ is bounded by a factor of $1/e$ \cite{Daulton2020qEHVI, Fisher1978}. This guarantee is myopic and applies per iteration, providing principled support for the greedy hypervolume-ranked batch rule.

\begin{algorithm}[t]
\caption{MOBO using Hypervolume-ranked Acquisition Extension (\textit{q}HAX-$\alpha$)}
\label{alg:qhax}
\begin{algorithmic}[1]
\State \textbf{Input:} Initial data $\mathcal{D}_0$, single-objective acquisition strategy $\alpha$, batch size $Q$
\For{$t = 0,1,2,\dots (T-1)$}
    \State Fit surrogate models $p(f^m \mid \mathcal{D}_t)$ for $m=1,\dots,M$
    \State Construct per-objective acquisitions $\alpha_t^{m}(\mathbf{x})$ and form vector acquisition $\boldsymbol{\alpha}_{t}(\mathbf{x})$
    \State Generate Pareto candidate set $\mathcal{C}^0_t$ by Pareto-optimizing $\boldsymbol{\alpha}_{t}(\mathbf{x})$
    \State Construct the denoised Pareto front $\mathcal{P}^0_t = \mathrm{ND}\big( \{\boldsymbol{\mu}_t(\mathbf{x}_i) \mid (\mathbf{x}_i,\cdot)\in\mathcal{D}_t\} \big)$
    \State Calculate reference point $r_t^m= \min [\mathcal{P}^0_t \cup \boldsymbol{\mu}_t(\mathcal{C}^0_t)]^m - 0.01$ for $m=1,\dots,M$
    \For{$q = 0,1,\dots (Q-1)$} 
        \State Construct utility $h^q_t(\mathbf{c}_t)=HV\big(\mathcal{P}^q_t \cup \{\boldsymbol{\mu}_{t}(\mathbf{c}_t)\}; \mathbf{r}_t\big)$
        \State Select $\mathbf{c}^{q+1}_{t+1} \leftarrow \arg\max_{\mathbf{c}_t \in \mathcal{C}^q_t} h^q_t(\mathbf{c}_{t})$, \quad (if $\mathcal{C}^q_t=\varnothing$, then $\mathbf{c}^{q+1}_{t+1} \leftarrow$  uniform sample in $\mathcal{X}$)
        \State Update candidate set $\mathcal{C}^{q+1}_t \leftarrow \mathcal{C}^q_t \backslash\{\mathbf{c}^{q+1}_{t+1}\}$ and Pareto-set $\mathcal{P}^{q+1}_t \leftarrow \text{ND}(\mathcal{P}^q_t \cup \{\boldsymbol{\mu}_{t}(\mathbf{c}^{q+1}_{t+1})\})$
    \EndFor
    \State Evaluate $\mathbf{f}(\mathbf{c}^{q}_{t+1})$ for $q=1,\dots,Q$ and update $\mathcal{D}_{t+1}$
\EndFor
\end{algorithmic}
\end{algorithm}

\subsection{Multi-Objective Hedge Strategy}
While \textit{q}HAX-$\alpha$ enables MOBO policies with distinct exploration behaviors, such as \textit{q}HAX-$UCB$ and \textit{q}HAX-$EI$, to be constructed modularly without bespoke derivations, the performance of the underlying acquisition functions remains problem dependent. Consequently, the most suitable acquisition behavior for an unknown black-box problem is difficult to identify \emph{a priori}. Motivated by this problem-dependent behavior, we extend the single-objective `GP--Hedge' strategy of Hoffman et al.~\cite{brochu2011portfolio} to the multi-objective setting. The proposed MO-Hedge approach adaptively selects among a portfolio of MOBO strategies with probabilities proportional to their historical Pareto-front contributions, mitigating acquisition mismatch online. Algorithm~\ref{alg:mo-hedge} summarizes the proposed approach. Although MO-Hedge pairs naturally with \textit{q}HAX-$\alpha$ portfolios, it is agnostic to the underlying acquisition formulation and can incorporate any batch-capable MOBO strategy.


\paragraph{Candidate Generation Per Acquisition:}
Hedge strategies require all acquisition functions to generate candidates independently based on the current posterior before selecting the winning candidates according to past performance. Since the acquisition functions operate independently, this step is easily parallelizable, and the additional computational cost is not a bottleneck for modern architectures~\cite{brochu2011portfolio}. Let $\{\alpha_1,\dots,\alpha_K\}$ denote a portfolio of $K$ acquisition functions for MOBO. At iteration $t$, each acquisition $\alpha_k$ proposes a batch of $Q$ candidates,
$
\mathcal{C}_t^k = \{\mathbf{x}_{t,1}^k,\dots,\mathbf{x}_{t,Q}^k\}.
$

Each acquisition maintains a \textit{history} of previously proposed candidates, $\mathcal{H}^k_t$, which enables retrospective evaluation of acquisition \textit{performance} under the current surrogate posterior. For the subsequent iteration, the history sets are updated as $\mathcal{H}^k_{t+1} \leftarrow \mathcal{C}_t^k \cup \mathcal{H}^k_t$.

\paragraph{Hypervolume-Based Rewards and Selection:}
Let $\mathcal{H}_t^k$ denote the set of candidates previously generated by acquisition $\alpha_k$. For each candidate $\mathbf{x} \in \mathcal{H}_t^k$, we compute the posterior-mean objective vector
$\boldsymbol{\mu}_t(\mathbf{x}) = \mathbb{E}[\mathbf{f}(\mathbf{x})\mid\mathcal{D}_t]$.
The reward (or \textit{gain}) for acquisition $\alpha_k$ is then defined via hypervolume
\[
g_t^k
=
HV\!\big(\{\boldsymbol{\mu}_t(\mathbf{x}) : \mathbf{x}\in \mathcal{H}_t^k\}; \mathbf{r}^{hedge}\big),
\]
where $HV(\cdot;\mathbf{r}^{hedge})$ denotes the hypervolume indicator with respect to a reference point $\mathbf{r}^{hedge}$. This choice rewards acquisitions that have historically proposed candidates contributing meaningfully to Pareto-front quality. Given gains $\{g_t^k\}_{k=1}^K$, acquisition selection probabilities are computed using an exponentiated-weights update \cite{brochu2010tutorial,492488}
\[
p_t^k
=
\frac{\exp(\eta_t g_t^k)}{\sum_{j=1}^K \exp(\eta_t g_t^j)}
\]
where $\eta_t > 0$ is a learning rate. Using selection probabilities $\{p_t^k\}$, a nominee $\mathbf{x}_{t,q}^k \in \mathcal{C}_t^k$ from the $k$-th acquisition function is selected with probability $p_t^k$ as the $q$-th MOBO evaluation candidate.

This multi-objective Hedge strategy preserves the adaptive acquisition-selection properties of classical GP--Hedge while introducing a Pareto-aware reward signal. Unlike the hypervolume-ranked selection used in \textit{q}HAX-$\alpha$, which resolves trade-offs among candidates generated within a single iteration, MO-Hedge adapts \emph{which} acquisition
function to trust over time based on historical performance.


\paragraph{Theoretical Remark -- Acquisition Selection Guarantee:}
Hoffman et al.~\cite{brochu2011portfolio} showed that, for GP--Hedge, the cumulative expert-selection regret relative to the best fixed acquisition in hindsight (conditioned on the portfolio-generated history) is sublinear, $\mathcal{O}(\sqrt{T})$, for an appropriate learning rate $\eta_t$~\cite{10.5555/1137817}. If MO-Hedge freezes prior rewards and defines incremental gains as $g_t^k = g_{t-1}^k + HV\big(\boldsymbol{\mu}_t(\mathcal{H}^k_t \setminus \mathcal{H}^k_{t-1}); \mathbf{r}^{hedge}\big)$, the standard Hedge regret bound applies under bounded-reward assumptions. In Algorithm~\ref{alg:mo-hedge}, however, gains are recalibrated \textit{retrospectively} over the entire history $\mathcal{H}^k_t$ using the updated posterior $\boldsymbol{\mu}_t$, which violates the standard Hedge assumptions. But this prevents stale early-surrogate rewards from accumulating. Ablations in Appendix~\ref{app:hedgeablations} confirm that this reward strategy yields a slight aggregate advantage over frozen-reward and one-step alternatives.


\begin{algorithm}[t]
\caption{Multi-Objective Hedge Acquisition Selection (MO-Hedge)}
\label{alg:mo-hedge}
\begin{algorithmic}[1]
\State \textbf{Input:} Portfolio $\{\alpha_1,\dots,\alpha_K\}$, histories $\mathcal{H}_0^k = \varnothing$, ref. point $\mathbf{r}^{hedge}$, batch size $Q$, learn rate $\eta_t$
\For{$k = 1,\dots,K$}
    \State Propose $Q$ acquisition nominees $\mathcal{C}_t^k = \{\mathbf{x}_{t,1}^k,\dots,\mathbf{x}_{t,Q}^k\}$
    \State Compute posterior-mean objectives set $\{\boldsymbol{\mu}_t(\mathbf{x})\}$ for all $\mathbf{x}\in\mathcal{H}_{t}^k$
    \State Calculate gains, $g_{t}^k=HV\!\big(\left\{\boldsymbol{\mu}_t(\mathbf{x}) : \mathbf{x}\in \mathcal{H}_t^k\right\}; \mathbf{r}^{{hedge}}\big)$, \quad (or $g_t^k = 0$ if $\mathcal{H}_t^k = \varnothing$)
    \State Update candidate history $\mathcal{H}^k_{t+1} \leftarrow \mathcal{C}_t^k \cup \mathcal{H}^k_t$
\EndFor
\State Compute selection probabilities $p_{t}^k=
\exp(\eta_t g_{t}^k)/{\sum_{j=1}^K \exp(\eta_t g_{t}^j)}$
\For{$q=1,\dots,Q$} 
\State Sample $k_q\sim\mathrm{Categorical} (p_t^1,\dots,p_t^K)$ and select $\mathbf{c}_{t+1}^{q} \leftarrow \mathbf{x}_{t,q}^{k_q}$ as candidate
\EndFor
\State \textbf{Output:} $Q$ evaluation candidates $\left\{\mathbf{c}^q_{t+1}\right\}_{q=1}^Q$
\end{algorithmic}
\end{algorithm}

\section{Benchmarking Details}

\paragraph{Acquisition Strategies:}
We empirically benchmark \textit{q}HAX-$\alpha$ and MO-Hedge on synthetic and application-derived optimization problems against state-of-the-art MOBO methods. We use the single-objective Upper Confidence Bound (UCB) and Expected Improvement (EI)  acquisition functions to instantiate \textit{q}HAX-$UCB$ and \textit{q}HAX-$EI$. We compare these methods against several baselines: the parallel noisy log variant of EHVI, \textit{q}LNEHVI (\textit{q}-Log Noisy Expected Hypervolume Improvement) \cite{Daulton2020qEHVI}; the scalarization-based \textit{q}LNParEGO \cite{Knowles2006ParEGO}; the hypervolume lookahead method \textit{q}HVKG (\textit{q}-Hypervolume Knowledge Gradient) \cite{pmlr-v202-daulton23a}; and the entropy-based method \textit{q}LB-MOJES (\textit{q}-Lower-Bound Multi-Objective Joint Entropy Search) \cite{Tu2022JES}. Implementation details are provided in Appendix~\ref{app:bench}.

We also benchmark two variants of MO-Hedge. The first, MO-Hedge 1, includes \textit{q}HAX-$UCB$, \textit{q}LNParEGO, and a random sampling strategy that returns a uniformly sampled point from the input space. MO-Hedge 1 is designed to test whether the hedge strategy can remain effective when its portfolio includes a weaker acquisition strategy and a naive random sampler. The second, MO-Hedge 2, includes the four strong acquisition strategies in our study: \textit{q}HAX-$UCB$, \textit{q}HAX-$EI$, \textit{q}HVKG, and \textit{q}LNEHVI. Together, these two variants allow us to assess the robustness and utility of MO-Hedge.

\paragraph{Benchmark Problems:} 
All algorithms are evaluated on twelve multi-objective benchmarks: nine established synthetic benchmarks and three application-derived benchmarks. The synthetic benchmarks comprise seven test functions and two engineering design problems. The test function benchmarks include ZDT2, ZDT4, and ZDT6 \cite{6787994}; DTLZ1, DTLZ2, and DTLZ3 \cite{1007032}; and Branin--Currin \cite{balandat2020botorch}. The engineering benchmarks are the unconstrained Welded Beam design problem \cite{COELLOCOELLO20021245} and the Vehicle Crash Safety design problem \cite{TANABE2020106078}. These test problems are widely used for benchmarking MOBO algorithms \cite{Daulton2020qEHVI, pmlr-v202-daulton23a, Tu2022JES}. We evaluate them across different input dimensions and numbers of objectives. DTLZ7 is included as an additional synthetic benchmark in Appendix~\ref{app:scaling}.

The algorithms are additionally evaluated on three heterogeneous application-derived benchmarks drawn from machine learning, aerodynamic design, and biochemical process engineering. The PD1 CIFAR-100 problem represents multi-objective hyperparameter optimization of Wide ResNet training for image classification on CIFAR-100 \cite{wang2024hyperbo}. The AirDbM problem considers aerodynamic shape optimization using a compact Design-by-Morphing representation of airfoil geometries \cite{10.1093/jcde/qwaf124,airfoils}. The Penicillin problem uses a process simulator to balance penicillin yield, fermentation time, and CO$_2$ emissions \cite{liang2021penicillin}. These application-derived problems provide valuable tests of the algorithms in more realistic settings. Detailed descriptions of all benchmark problems are provided in Appendix~\ref{app:bench}.

\paragraph{Results:}
Each benchmark problem was evaluated over 20 independent runs per acquisition strategy, initialized from a shared 30-point design $\mathcal{D}_0$ to ensure a fair comparison and run for 75 additional function evaluations. Observation noise was added independently to each objective, with standard deviation set to \(1\%\) of the absolute mean objective value over $\mathcal{D}_0$.

\begin{figure}[tb]
\includegraphics[width=0.925\linewidth]{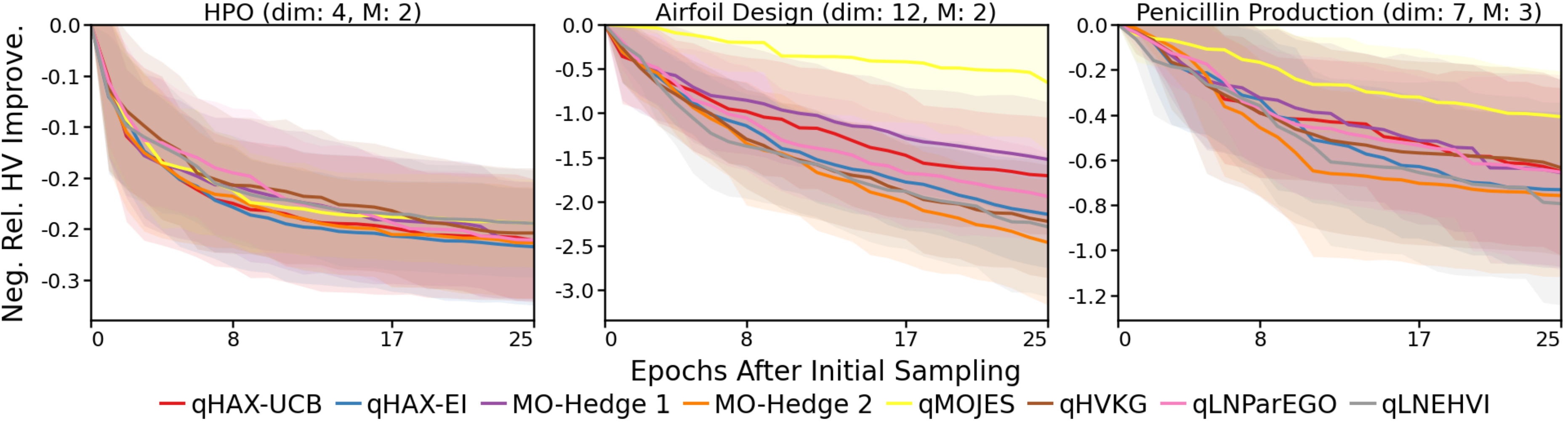}
\centering
\vspace{-2mm}
\caption{Neg. relative $HV$ improvement for batch optimization (\(Q=3\)) across application-derived benchmarks. Lower values are better.
Mean (solid line)  \(\pm 1\) standard deviation (shaded)  over 20 runs.}
\label{fig:real}
\vspace{-3.5mm}
\end{figure}

For the nine synthetic benchmarks, we evaluate performance in both serial (\(Q=1, T=75~\mathrm{epochs}\)) and parallel batch (\(Q=3,  T=25~\mathrm{epochs}\)) settings using hypervolume (\(HV\)) and inverted generational distance (\(IGD\)). For external \(HV\) evaluation, the reference point $\mathbf{r}^{{eval}}$ is determined prior to optimization from $\mathcal{D}_0$ and held fixed across all algorithms and epochs within each paired run. Results are reported using the log-normalized indicator gap, \(\log\left(\left| I(\mathcal{P}_{\mathrm{opt}}) - I(\mathcal{P}_{\mathrm{curr}}) \right| \big/ \left| I(\mathcal{P}_{\mathrm{opt}}) - I(\mathcal{P}_{\mathrm{init}}) \right|\right)\), where \(\mathcal{P}_{\mathrm{opt}}\), \(\mathcal{P}_{\mathrm{curr}}\), and \(\mathcal{P}_{\mathrm{init}}\) denote the denoised nondominated Pareto fronts from a high-fidelity NSGA-II reference solution, the current dataset, and $\mathcal{D}_0$, respectively. Figure~\ref{fig:q3hv} shows the mean log-normalized \(HV\) gap (\(\pm 1\) standard deviation shaded) over 20 runs for \(Q=3\); corresponding \(IGD\) and serial (\(Q=1\)) results are reported in Appendix~\ref{app:results}, and scaling studies across higher input dimensions and objective counts are provided in Appendix~\ref{app:scaling}. The three application-derived benchmarks are only evaluated in the batch setting (\(Q=3\)). Since accurate reference Pareto fronts are unavailable for these problems, Figure~\ref{fig:real} instead reports the negative relative hypervolume improvement, \(\left({HV}(\mathcal{P}_{\mathrm{init}};\mathbf{r}^{{eval}}) - {HV}(\mathcal{P}_{curr};\mathbf{r}^{{eval}})\right) \big/ {HV}(\mathcal{P}_{\mathrm{init}};\mathbf{r}^{{eval}})\), where \(\mathcal{P}_{\mathrm{init}}\) and \(\mathcal{P}_{curr}\) denote the nondominated objective sets at initialization and iteration \(t\), respectively. Starting at zero, the metric becomes increasingly negative as hypervolume improves.

\section{Discussion}
\begin{figure}[tb]
\includegraphics[width=0.925\linewidth]{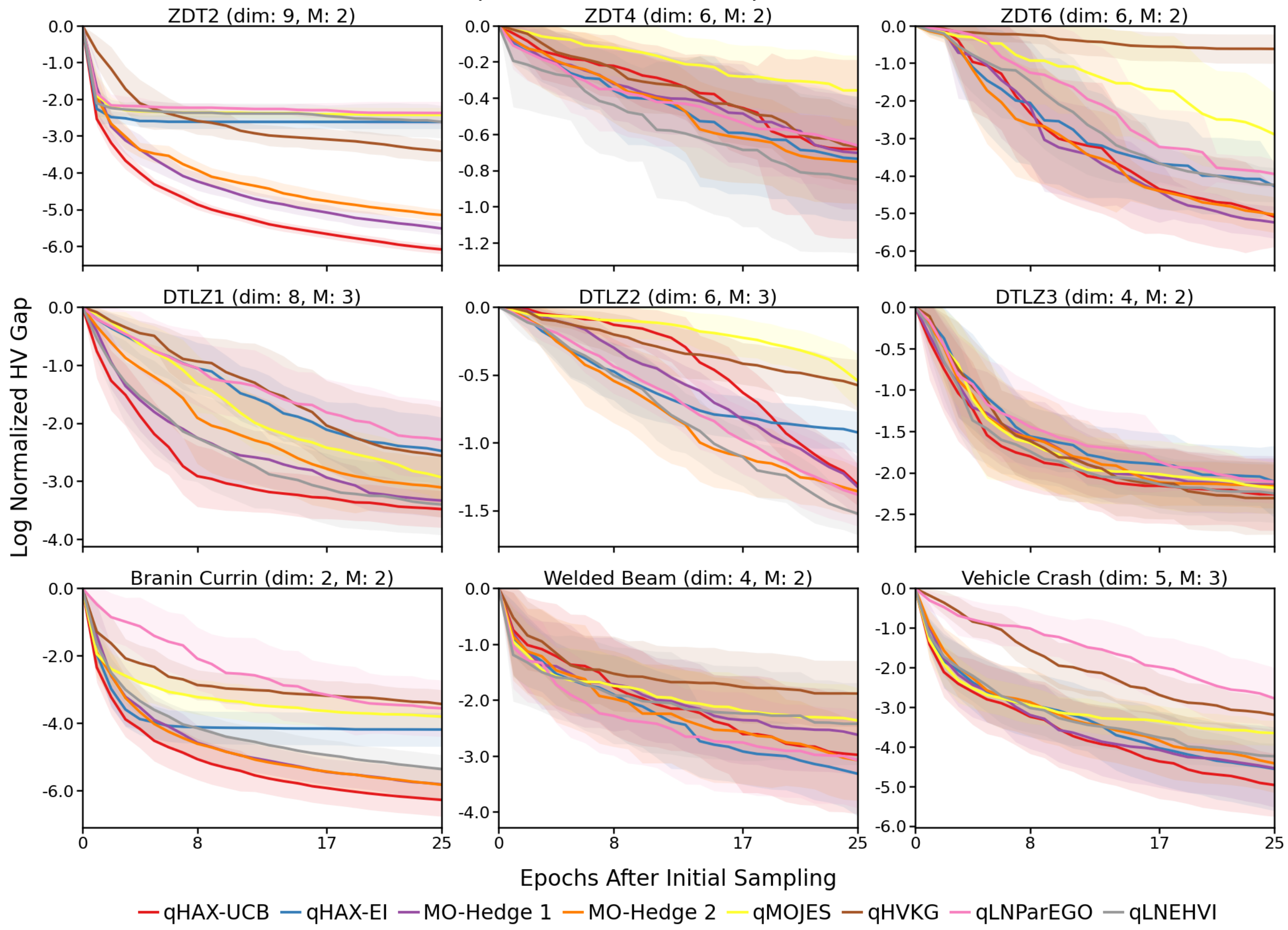}
\centering
\vspace{-2mm}
\caption{Log-normalized hypervolume gap for batch optimization (\(Q=3\)) across synthetic benchmarks. Lower values are better.
Mean (solid line)  \(\pm 1\) standard deviation (shaded)  over 20 runs.}
\label{fig:q3hv}
\vspace{-3.5mm}
\end{figure}

\paragraph{\textit{q}HAX-$\boldsymbol{\alpha}$ Performance:}
Across the benchmark suite, \textit{q}HAX-$\alpha$ performs competitively with, and in several cases outperforms, more specialized state-of-the-art MOBO baselines. On the synthetic benchmarks, the \textit{q}HAX-$\alpha$ variants converge rapidly toward the reference Pareto front. In particular, \textit{q}HAX-$UCB$ exhibits strong and consistent performance across most benchmark problems, achieving the best or near-best final hypervolume gap on several tasks. Log-normalized IGD results (Appendix~\ref{app:results}) show similar trends, indicating good coverage of the Pareto front. The performance of \textit{q}HAX-$EI$ is more problem dependent. While it remains competitive on several benchmarks, particularly the engineering design and application-derived problems, it is generally less robust than \textit{q}HAX-$UCB$ across the synthetic benchmarks. This behavior is consistent with the known tendency of EI-based acquisitions to become increasingly exploitative once the surrogate model identifies promising regions, compared with the more sustained exploration encouraged by UCB. The application-derived results also show that this behavior can be advantageous on some problems, with \textit{q}HAX-$EI$ achieving a lower final mean value than \textit{q}HAX-$UCB$ on all three problems.

Several baseline algorithms
remain competitive on specific benchmark problems, but none consistently outperforms the \textit{q}HAX-$\alpha$ variants across the complete benchmark suite. This strong overall performance is maintained as the input dimension and number of objectives increase, as shown in Appendix~\ref{app:scaling}. Importantly, no current state-of-the-art MOBO strategy dominates uniformly across the benchmark suite. The strong performance of \textit{q}HAX-$UCB$ is therefore notable, as it is achieved without constructing a bespoke multi-objective acquisition function. These results support the central hypothesis of this work: a lightweight, Pareto-aware extension of classical single-objective acquisition functions can yield consistently strong performance across a wide range of multi-objective optimization problems. Component ablations in Appendix~\ref{app:acqablations} isolate the two stages of \textit{q}HAX-$\alpha$, showing that acquisition-space Pareto candidate generation is the primary driver of exploration and performance (which cannot be recovered by hypervolume ranking over random candidates), while posterior-mean hypervolume ranking provides the most consistent candidate selection rule compared 
with random, distance-based, and summed-acquisition alternatives.

\paragraph{MO-Hedge Performance:}
As shown in Fig.~\ref{fig:q3hv}, the relative performance of individual MOBO acquisition strategies varies substantially across benchmarks. These results reinforce our central motivation for MO-Hedge: no single acquisition strategy performs uniformly well across all benchmark problems, underscoring the need for adaptive acquisition selection rather than committing \emph{a priori} 
to a fixed strategy for a black-box function with unknown structure.
MO-Hedge directly addresses this challenge by learning which acquisition strategy to trust as optimization progresses. Both MO-Hedge variants demonstrate robust performance across the benchmark suite. In particular, MO-Hedge 1 exhibits consistently strong performance despite containing a deliberately mixed-strength portfolio comprising \textit{q}HAX-$UCB$, \textit{q}LNParEGO, and a random sampling strategy. Across benchmarks, MO-Hedge 1 reliably tracks the performance of its strongest constituent, indicating that the hedge mechanism effectively downweights ineffective strategies while concentrating probability mass on more informative strategies. MO-Hedge 2, which draws from a stronger portfolio of acquisition strategies, performs consistently well across all benchmark problems. While individual acquisitions in this portfolio may underperform, the adaptive selection mechanism enables MO-Hedge 2 to maintain stable and competitive performance. An ablation study of the different reward strategies for MO-Hedge is presented in Appendix~\ref{app:hedgeablations}. Additionally, 
the appendix also presents the evolution of MO-Hedge's acquisition-selection probabilities over the course of optimization, illustrating the adaptive behavior of the hedge mechanism.

\paragraph{Runtime Analysis:}
Table~\ref{tab:runtime_comparison_2} reports the average wall-clock time required for each method to complete a single optimization run. Across all benchmarks, \textit{q}HAX-$\alpha$ exhibits low and stable runtime, with both \textit{q}HAX-$UCB$ and \textit{q}HAX-$EI$ showing little variation with problem dimension or the number of objectives. In contrast, more specialized MOBO baselines, particularly the entropy-based \textit{q}MOJES, incur substantially higher computational costs, often by an order of magnitude. Scalarization- and noisy hypervolume-based methods fall in between but still scale less favourably than \textit{q}HAX-$\alpha$. 
\begin{wraptable}{r}{0.638\textwidth}
\centering
\scriptsize
\setlength{\tabcolsep}{1.9pt}
\begin{tabular}{lcccccc}
\toprule
Test Function & \textit{q}HAX-$UCB$ & \textit{q}HAX-$EI$ & \textit{q}MOJES & \textit{q}HVKG & \textit{q}LNParEGO & \textit{q}LNEHVI \\
\midrule
ZDT2 & $230.41$ & $226.79$ & $1834.78$ & $268.04$ & $454.05$ & $673.54$ \\
ZDT4 & $214.38$ & $219.12$ & $2817.48$ & $296.25$ & $506.43$ & $680.57$ \\
ZDT6 & $221.89$ & $214.73$ & $8092.44$ & $321.96$ & $799.51$ & $953.11$ \\
DTLZ1 & $254.31$ & $246.49$ & $4636.83$ & $424.22$ & $372.96$ & $711.56$ \\
DTLZ2 & $243.24$ & $243.02$ & $4089.77$ & $390.18$ & $352.85$ & $886.75$ \\
DTLZ3 & $220.59$ & $218.35$ & $2943.21$ & $229.57$ & $507.18$ & $506.37$ \\
Branin--Currin & $246.04$ & $242.39$ & $2319.57$ & $211.54$ & $371.63$ & $467.71$ \\
Welded Beam & $226.48$ & $223.13$ & $3394.43$ & $272.27$ & $434.10$ & $519.02$ \\
Vehicle Crash & $242.42$ & $235.32$ & $4111.98$ & $339.72$ & $353.80$ & $753.87$ \\
\midrule
HPO & $257.74$ & $256.38$ & $1825.39$ & $293.23$ & $146.69$ & $293.94$ \\
Airfoil Design & $742.84$ & $941.94$ & $3084.95$ & $809.60$ & $631.17$ & $750.44$ \\
Penicillin Prod. & $269.58$ & $265.67$ & $3418.75$ & $499.66$ & $186.79$ & $1180.26$ \\
\bottomrule
\end{tabular}
\caption{Mean runtime in seconds for each algorithm and benchmark over 20 independent runs (batch mode: $Q=3$).}
\label{tab:runtime_comparison_2}
\end{wraptable}
The efficiency of \textit{q}HAX-$\alpha$ stems from its modular design: the dominant acquisition-related cost arises from the inner Pareto-front optimization used to generate acquisition candidates, which remains largely invariant to batch size and output count. The higher runtime observed for the Airfoil Design benchmark is dominated by the solver cost rather than the cost of the acquisition strategy itself. We do not report runtimes for MO-Hedge variants here, as their runtimes are approximately equal to the sum of the runtimes of their component acquisition strategies.

\paragraph{Limitations:}
A limitation of \textit{q}HAX-$\alpha$ is its reliance on an inner evolutionary optimizer (NSGA-II) to generate Pareto-diverse candidates, which does not directly utilize gradient-based optimization \cite{balandat2020botorch}, though hybrid evolutionary--gradient refinement can alleviate this \cite{10.1007/3-540-44719-9_27, Knowles2000-ua}. Additionally, during batch selection ($Q > 1$), \textit{q}HAX-$\alpha$ holds the surrogate posterior fixed rather than conditioning on fantasy observations; incorporating cheap rank-one fantasy updates to further promote within-batch diversity is a promising extension. Our empirical scalability claims are restricted to the tested regime of $d \le 20$ inputs and $M \le 4$ objectives. Finally, while automated rules define the internal reference points without problem-specific tuning, we have not conducted reference-point ablation studies.

\section{Conclusion}
This work introduced \textit{q}HAX-$\alpha$, a lightweight framework for multi-objective Bayesian optimization that extends standard single-objective acquisition functions through Pareto-diverse candidate generation and hypervolume-based selection. Across a diverse set of synthetic, engineering and application-derived benchmarks, \textit{q}HAX-$\alpha$, particularly \textit{q}HAX-$UCB$, achieved competitive or superior performance compared to state-of-the-art MOBO methods, while maintaining lower and more predictable computational cost. We further proposed MO-Hedge, a multi-objective portfolio strategy that adaptively selects acquisition functions based on historical Pareto-front contributions. Empirical results demonstrated that MO-Hedge is robust to acquisition mismatch and performs consistently well
across problem classes. Overall, our work highlights the practical value of modular, Pareto-aware extensions of classical single-objective acquisition strategies, and underscores adaptive acquisition selection using a hedge strategy as a key ingredient for robust multi-objective Bayesian optimization.

\begin{ack}
The authors acknowledge the use of the IRIDIS High Performance Computing Facility and associated support services at the University of Southampton in the completion of this work. We also thank Sangjoon Lee of Nanyang Technological University for assistance with the AirDbM benchmark. This research received no external funding. The authors declare no conflicts of interests.
\end{ack}
\bibliographystyle{unsrtnat}
\bibliography{ReferencesClean}

\newpage
\appendix

\section{Classical Single-Objective Acquisition Functions And Hedge Strategies}
\label{app:SO}

Single-objective (SO) acquisition functions translate the surrogate posterior's predicted mean and uncertainty into black-box function evaluation decisions. While the surrogate posterior provides a probabilistic description of the unknown objective function, the acquisition function determines how this information is acted upon, balancing exploitation of promising regions against exploration of uncertain ones. Although acquisition functions are often treated as modular components, their behavior can differ substantially across problem classes, and no single choice is uniformly optimal. Below, we provide a brief overview of the classical SO acquisition functions and portfolio-based hedge strategies employed in this work.

\subsection{Single-Objective Acquisition Functions}

Consider maximization of a black-box objective $f:\mathcal{X}\rightarrow\mathbb{R}$.
Given observations $\mathcal{D}_t=\{(\mathbf{x}_i,y_i)\}_{i=1}^{n_t}$, a Gaussian-process surrogate induces
a posterior predictive distribution
\[
p(f(\mathbf{x})\mid\mathcal{D}_t)
=
\mathcal{N}\big(\mu_t(\mathbf{x}),\,\sigma_t^2(\mathbf{x})\big)
\]
SO acquisition functions define a utility $\alpha_t(\mathbf{x})$ that is optimized to
select the next evaluation point.


\paragraph{Expected Improvement (EI):}
Expected Improvement incorporates the magnitude of improvement rather than only its
probability:
\[
{EI}_t(\mathbf{x})
=
\mathbb{E}\!\left[\max\big(f(\mathbf{x})-f_t^+-\xi,\,0\big)\mid\mathcal{D}_t\right],
\]
where $f_t^+$ denotes the best observed value. Under a GP posterior, EI admits a closed-form expression,
\[
{EI}_t(\mathbf{x})
=
(\mu_t(\mathbf{x})-f_t^+-\xi)\,\Phi(z)
+
\sigma_t(\mathbf{x})\,\phi(z),
\qquad
z=\frac{\mu_t(\mathbf{x})-f_t^+-\xi}{\sigma_t(\mathbf{x})},
\]
where $\phi(\cdot)$ and $\Phi(\cdot)$ denote the standard normal PDF and CDF, respectively.
EI provides a balanced exploration--exploitation trade-off and is widely used
in practice.

\paragraph{Upper Confidence Bound (UCB):}
Upper Confidence Bound acquisition functions adopt an explicitly optimistic strategy:
\[
{UCB}_t(\mathbf{x})
=
\mu_t(\mathbf{x}) + \beta_t^{1/2}\,\sigma_t(\mathbf{x}),
\]
where $\beta_t$ controls the degree of exploration. With an appropriate schedule for
$\beta_t$, GP-UCB has been proven to have sublinear cumulative regret guarantees under suitable assumptions \cite{Srinivas2010GPUCB}.

\subsection{SO Hedge Strategies for Acquisition Selection}
The practical performance of Bayesian optimization can depend strongly on the choice of acquisition function. Improvement-based criteria, confidence-bound methods, and sampling-based policies encode different exploration--exploitation biases, and their relative performance may vary with the smoothness, noise level, dimensionality, and multi-modality of the objective. This observation has motivated portfolio approaches that avoid committing to a single acquisition function \emph{a priori}.

The theoretical foundation for these approaches is closely related to online learning and multi-armed bandit algorithms, where a learner repeatedly selects among a set of arms and seeks to compete with the best arm in hindsight. In particular, adversarial bandit methods such as those of Auer et al.~\cite{492488} provide sublinear regret guarantees for exponentially weighted selection rules, establishing a principled basis for adaptive selection among competing strategies. In the BO setting, Hoffman et al.~\cite{brochu2011portfolio} introduced GP-Hedge, which treats acquisition functions as arms in a portfolio. At iteration $t$, each acquisition $\alpha^k$ proposes a candidate and an acquisition index is sampled according to the exponentiated-weights distribution,
\[
p_{t}^k=\frac{\exp(\eta g_{t-1}^k)}{\sum_{j=1}^{K}\exp(\eta g_{t-1}^j)}
\]
where $g_{t-1}^k$ denotes the cumulative \textit{gain} of acquisition $\alpha^k$ and $\eta>0$ is a learning rate. After the new evaluation and updating the surrogate, gains are updated using a reward assigned to each acquisition. Under the standard hedge assumptions, this yields a sublinear cumulative regret guarantee, $\mathcal{O}(\sqrt{T})$, with respect to the best fixed acquisition function in hindsight. Importantly, this regret is defined over acquisition selection, and not directly over optimization of the unknown objective.


\section{Algorithm Implementation and Benchmarking Details}
\label{app:bench}

\subsection{Algorithm Implementation in BoTorch}
This appendix summarizes the implementation details for all acquisition strategies used in our empirical evaluation. All acquisition functions are implemented in BoTorch~\cite{balandat2020botorch} and optimized from the same initial design set over identical bounded design domains to ensure fair comparison. All model-based acquisition strategies employ BoTorch's \texttt{ModelListGP} together with \texttt{SumMarginalLogLikelihood} for surrogate modeling. BoTorch provides native implementations of the baseline multi-objective methods considered in this work, including hypervolume-based acquisition functions such as \textit{q}LNEHVI~\cite{Daulton2020qEHVI}, scalarization-based approaches such as \textit{q}LNParEGO~\cite{Knowles2006ParEGO}, and the information-theoretic JES acquisition~\cite{Tu2022JES}. Baseline hyperparameters are chosen to closely match recommended default settings while maintaining consistent optimization budgets across methods. Our proposed \textit{q}HAX-$\alpha$ and MO-Hedge strategies are likewise implemented in BoTorch, with Pareto-front generation in acquisition space performed using \texttt{pymoo}'s NSGA-II~\cite{pymoo}. Across all synthetic benchmarks, application-derived problems, scaling studies, and ablation experiments, all hyperparameters for both the proposed methods and the baseline algorithms are held strictly fixed without problem-specific tuning. The complete implementation can be accessed open-source from \url{https://github.com/h7ris/MOBO-qHAX-MOHedge}, and full implementation details are provided in this section to support reproducibility.

For all methods, the bounded search domain $[\boldsymbol{\ell}, \mathbf{u}] \subset \mathbb{R}^d$ is normalized to the unit hypercube $\mathbf{x} \in [0,1]^d$, and observed objectives are standardized to zero mean and unit variance prior to fitting the GP surrogates at each iteration. For all BoTorch baselines, candidate generation is performed over $\mathbf{x} \in [0,1]^d$ using BoTorch's \texttt{optimize\_acqf} routine with $\texttt{num\_restarts}=10$, $\texttt{raw\_samples}=500$, $\texttt{batch\_limit}=5$, and $\texttt{maxiter}=100$, unless stated otherwise. 



\paragraph{\textit{q}HAX-$\alpha$ and MO-Hedge:}
To demonstrate that standard, off-the-shelf single-objective acquisitions can be extended directly without log-transformed or noisy multi-objective re-derivations, \textit{q}HAX-EI uses the classical analytical Expected Improvement formulation with $\xi = 0.01$, and \textit{q}HAX-UCB uses the GP-UCB exploration schedule~\cite{Srinivas2010GPUCB} under the parameterization of Hoffman et al.~\cite{brochu2011portfolio},
\[
\beta_t^{1/2} = \sqrt{\nu \left[ 2 \log\!\left( \frac{n_t^{d/2 + 2}\pi^2}{3\delta} \right) \right]},
\]
with $\nu = 0.2$ and $\delta = 0.1$, where $n_t = |\mathcal{D}_t|$ is the number of observations at iteration $t$. To generate the candidate set $\mathcal{C}_t^0$, NSGA-II~\cite{pymoo} is configured with a population size of 300 and 450 generations, with duplicate elimination enabled and default crossover and mutation operators, retaining the complete final nondominated population. In MO-Hedge, portfolio selection probabilities are updated using the time-varying learning rate schedule of Hoffman et al.~\cite{brochu2011portfolio},
\[
\eta_t = \sqrt{\frac{8 \log K}{n_t}},
\]
where $K$ denotes the portfolio size. Historical hypervolume gains $g_t^k$ are evaluated in standardized posterior-mean objective space with respect to the fixed reference point $\mathbf{r}^{\mathrm{hedge}} = (-3, \dots, -3) \in \mathbb{R}^M$ (three standard deviations below the empirical objective mean) without additional reward normalization.

\paragraph{\textit{q}LNEHVI:}
The \textit{q}-Log Noisy Expected Hypervolume Improvement variant uses BoTorch’s \texttt{qLogNoisyExpectedHypervolumeImprovement}. At each iteration \(t\), posterior‑mean predictions are computed at the evaluated design points and used to construct a fast nondominated partitioning of objective space. A batch of candidate points is then obtained by jointly maximizing this acquisition with BoTorch’s \texttt{optimize\_acqf} routine. In the parallel batch setting ($Q=3$), \textit{q}LNEHVI is optimized jointly over all $Q$ candidates (\texttt{sequential=False}) to target their exact multi-point acquisition objectives and account for within-batch interactions, which explains its higher runtime in batch mode (Table~1) relative to sequential mode ($Q=1$, Table~3).

\paragraph{\textit{q}LNParEGO:}
The \textit{q}LNParEGO baseline implements a parallel ParEGO‑style scalarization strategy using a BoTorch wrapper around Chebyshev scalarization and log‑noisy improvement. In ParEGO‑style methods, a randomly sampled scalarization converts the vector‑valued objective into a scalar objective, after which standard single‑objective acquisition optimization can be applied. In our implementation, the acquisition is instantiated with the current model and training inputs and then optimized jointly over a batch of size \(q\) using \texttt{optimize\_acqf}. Similar to \textit{q}LNEHVI, in the parallel batch setting ($Q=3$), \textit{q}LNParEGO is optimized jointly over all $Q$ candidates (\texttt{sequential=False}). 

\paragraph{\textit{q}HVKG:}
The \textit{q}-Hypervolume Knowledge Gradient baseline uses BoTorch’s \texttt{qHypervolumeKnowledgeGradient}. HVKG is a one‑step lookahead acquisition function that maximizes the expected increase in the hypervolume of a finite posterior‑mean Pareto approximation. We adopt a deterministic sample‑average approximation (SAA) formulation with fixed Sobol‑QMC base samples for generating fantasies. Specifically, we use \(\texttt{num\_fantasies}=10\) and \(\texttt{num\_pareto}=10\), and construct a \texttt{ListSampler}. The acquisition is optimized using \texttt{optimize\_acqf} with a single start.

HVKG candidate generation can occasionally fail due to numerical optimization issues. In such cases, we employ a fallback procedure consisting of a conservative Sobol search over candidate batches: 500 Sobol batches are drawn uniformly from the design space, evaluated under the HVKG acquisition, and the best batch is returned. This fallback is used solely as a numerical safeguard and is not part of the primary HVKG baseline.

\paragraph{\textit{q}MOJES:}
The qLB‑MOJES baseline implements the lower‑bound \textit{q}-batch variant of Multi‑Objective Joint Entropy Search (MOJES). Following the default implementation, Pareto‑set samples are generated using Matheron path models. We use 20 Pareto‑set samples, 40 Pareto points per sample, 8 random scalarizations per path, 1000 entropy‑estimation samples, and 500 Sobol samples for additional candidate coverage. For each sampled path, candidate optima are generated by optimizing random scalarizations with Dirichlet‑distributed weights, and Sobol samples are included to improve coverage of the design space. Given the sampled Pareto fronts, we construct the dominated‑space hypercell decomposition using  \texttt{DominatedPartitioning}. For \(M>2\), where batched dominated partitioning is not available in the BoTorch version used, a separate partitioning is constructed for each Pareto front sample and combined using \texttt{BoxDecompositionList}. The resulting acquisition is optimized jointly over \(q\) candidates using \texttt{optimize\_acqf}.

\subsection{Computational Resource Details}
All benchmarks were run on a CPU-based high-performance computing cluster with AMD EPYC 9654 96-core processors. To ensure fair comparison, each experimental run, where all algorithms were evaluated sequentially starting from the same initial dataset, was executed with an allocation of 12 CPU cores and 160~GB of shared memory shared across parallel processes.

\subsection{Multi-Objective Benchmark Problem Details}
This appendix details the benchmark optimization problems used in our empirical evaluation. All optimization problems are stated directly in maximization form, consistent with the main paper. Unless otherwise stated, noisy observations for each objective $m \in \{1,\dots,M\}$ are generated as
\[
y^m(\mathbf{x}) = f^m(\mathbf{x}) + \varepsilon^m,
\qquad
\varepsilon^m \sim \mathcal{N}\big(0,(\sigma_n^m)^2\big),
\]
with independent Gaussian noise across objectives and standard deviation $\sigma_n^m$ set to $1\%$ of the absolute mean objective value over the initial design $\mathcal{D}_0$.

The synthetic benchmark test functions include ZDT2, ZDT4, and ZDT6 \cite{6787994}; DTLZ1, DTLZ2, DTLZ3, and DTLZ7 \cite{1007032}; and Branin--Currin \cite{balandat2020botorch}. The engineering design problems are the unconstrained Welded Beam problem \cite{COELLOCOELLO20021245} and the Vehicle Crash Safety problem \cite{TANABE2020106078}. These test problems are widely used for benchmarking MOBO algorithms \cite{Daulton2020qEHVI, pmlr-v202-daulton23a, Tu2022JES}. In addition, the suite includes three application-derived benchmarks: Wide ResNet hyperparameter optimization (HPO) on CIFAR-100 \cite{wang2024hyperbo}, aerodynamic shape optimization using AirDbM \cite{10.1093/jcde/qwaf124,airfoils}, and the Penicillin production problem \cite{liang2021penicillin}. The input dimensions and number of objectives for the benchmark suite are summarized in Table~\ref{tab:test-functions}.

\paragraph{ZDT2:}
This is a two-objective benchmark from the Zitzler--Deb--Thiele (ZDT) test suite \cite{6787994}, designed
to test convergence to a smooth, non-convex Pareto front. We use a
\(d\)-dimensional decision vector \(\mathbf{x}\in[0,1]^d\). The maximization objectives are
\[
f^1(\mathbf{x}) = -x_1
\]
\[
g(\mathbf{x})
=
1+\frac{9}{d-1}\sum_{i=2}^{d}x_i
\]
\[
f^2(\mathbf{x})
=
-
g(\mathbf{x})
\left[
1-\left(\frac{x_1}{g(\mathbf{x})}\right)^2
\right]
\]
The Pareto-optimal set is obtained when \(x_i=0\) for all \(i=2,\dots,d\).

\paragraph{ZDT4:}
A two-objective benchmark with a highly multi-modal distance function. The input bounds are
$
x_1\in[0,1],~
x_i\in[-5,5] ~\text{for } i=2,\dots,d.
$
The maximization objectives are
\[
f^1(\mathbf{x})=-x_1
\]
\[
g(\mathbf{x})
=
1+10(d-1)
+
\sum_{i=2}^{d}
\left[
x_i^2-10\cos(4\pi x_i)
\right]
\]
\[
f^2(\mathbf{x})
=
-
g(\mathbf{x})
\left[
1-\sqrt{\frac{x_1}{g(\mathbf{x})}}
\right]
\]
The Pareto-optimal set is obtained when \(x_i=0\) for all \(i=2,\dots,d\).

\paragraph{ZDT6:}
ZDT6 is a two-objective benchmark with a non-uniform Pareto front and a nonlinear first objective. The domain is
\(\mathbf{x}\in[0,1]^d\). Let
\[
s(\mathbf{x})
=
1-\exp(-4x_1)\sin^6(6\pi x_1)
\]
\[
g(\mathbf{x})
=
1+9
\left(
\frac{1}{d-1}\sum_{i=2}^{d}x_i
\right)^{1/4}
\]
The maximization objectives are
\[
f^1(\mathbf{x})=-s(\mathbf{x})
\]
\[
f^2(\mathbf{x})
=
-
g(\mathbf{x})
\left[
1-\left(\frac{s(\mathbf{x})}{g(\mathbf{x})}\right)^2
\right]
\]
The Pareto-optimal set is obtained when \(x_i=0\) for all \(i=2,\dots,d\).

\paragraph{DTLZ1:}
This is a scalable many-objective benchmark from the Deb--Thiele--Laumanns--Zitzler (DTLZ) suite \cite{1007032}. It has a linear Pareto front and a multi-modal distance function with many local Pareto fronts. Let \(\mathbf{x}\in[0,1]^d\), let \(M\) denote the number
of objectives, and define
\[
k=d-M+1,
\qquad
\mathbf{x}_M=(x_M,\dots,x_d)
\]
The distance function is
\[
g(\mathbf{x}_M)
=
100\left[
k+
\sum_{x_i\in\mathbf{x}_M}
\left(
(x_i-0.5)^2-\cos(20\pi(x_i-0.5))
\right)
\right]
\]
For objective index \(m=1,\dots,M\), define
\[
\psi^m(\mathbf{x})
=
\begin{cases}
1, & m=1\\
1-x_{M-m+1}, & m>1
\end{cases}
\]
The maximization objectives are
\[
f^m(\mathbf{x})
=
-
\frac{1}{2}(1+g(\mathbf{x}_M))
\left(
\prod_{i=1}^{M-m}x_i
\right)
\psi^m(\mathbf{x}),
\qquad
m=1,\dots,M
\]
The global Pareto-optimal set satisfies \(x_i=0.5\) for all \(x_i\in\mathbf{x}_M\), yielding
\(g(\mathbf{x}_M)=0\).

\paragraph{DTLZ2:}
A scalable benchmark with a spherical Pareto front and a uni-modal distance function. Let \(\mathbf{x}\in[0,1]^d\), let \(M\) denote the number of objectives, and let
$\mathbf{x}_M=(x_M,\dots,x_d)$.
The distance function is
\[
g(\mathbf{x}_M)
=
\sum_{x_i\in\mathbf{x}_M}(x_i-0.5)^2
\]
For \(m=1,\dots,M\), define
\[
\psi^m(\mathbf{x})
=
\begin{cases}
1, & m=1\\
\sin\left(\dfrac{\pi x_{M-m+1}}{2}\right), & m>1
\end{cases}
\]
The maximization objectives are
\[
f^m(\mathbf{x})
=
-
(1+g(\mathbf{x}_M))
\left(
\prod_{i=1}^{M-m}
\cos\left(\frac{\pi x_i}{2}\right)
\right)
\psi^m(\mathbf{x}),
\qquad
m=1,\dots,M
\]
The Pareto-optimal set satisfies \(x_i=0.5\) for all \(x_i\in\mathbf{x}_M\), giving
\(g(\mathbf{x}_M)=0\).

\paragraph{DTLZ3:} A scalable benchmark from the DTLZ suite that combines the spherical objective mapping of DTLZ2 with the multi-modal distance function of DTLZ1. The domain is \(\mathbf{x}\in[0,1]^d\). Let
\[
k=d-M+1,
\qquad
\mathbf{x}_M=(x_M,\dots,x_d)
\]
The distance function is
\[
g(\mathbf{x}_M)
=
100\left[
k+
\sum_{x_i\in\mathbf{x}_M}
\left(
(x_i-0.5)^2-\cos(20\pi(x_i-0.5))
\right)
\right]
\]
Using the same spherical mapping as DTLZ2, define
\[
\psi^m(\mathbf{x})
=
\begin{cases}
1, & m=1\\
\sin\left(\dfrac{\pi x_{M-m+1}}{2}\right), & m>1
\end{cases}
\]
The maximization objectives are
\[
f^m(\mathbf{x})
=
-
(1+g(\mathbf{x}_M))
\left(
\prod_{i=1}^{M-m}
\cos\left(\frac{\pi x_i}{2}\right)
\right)
\psi^m(\mathbf{x})
\qquad
m=1,\dots,M
\]
The global Pareto-optimal set is obtained when \(x_i=0.5\) for all distance variables
\(x_i\in\mathbf{x}_M\).

\paragraph{DTLZ7:}
A scalable benchmark from the DTLZ suite with $2^{M-1}$ separated Pareto-optimal regions, included in our scaling studies (Appendix~\ref{app:scaling}) to evaluate optimization efficacy on a disconnected Pareto front. Let \(\mathbf{x}\in[0,1]^d\), let \(M\) denote the number of objectives, and define
\[
k=d-M+1,
\qquad
\mathbf{x}_M=(x_M,\dots,x_d).
\]
The distance function is
\[
g(\mathbf{x}_M)
=
1 + \frac{9}{k} \sum_{x_i\in\mathbf{x}_M} x_i,
\]
and the shape function is defined over the first \(M-1\) decision variables as
\[
h(\mathbf{x})
=
M - \sum_{i=1}^{M-1}
\left[
\frac{x_i}{1+g(\mathbf{x}_M)}
\big(1+\sin(3\pi x_i)\big)
\right].
\]
The maximization objectives are
\[
f^m(\mathbf{x})
=
\begin{cases}
-x_m, & m=1,\dots,M-1,\\[4pt]
-\big(1+g(\mathbf{x}_M)\big)\,h(\mathbf{x}), & m=M.
\end{cases}
\]
The global Pareto-optimal set satisfies \(x_i=0\) for all \(x_i\in\mathbf{x}_M\), yielding \(g(\mathbf{x}_M)=1\).

\paragraph{Branin--Currin:}
The Branin--Currin benchmark is a two-objective problem that combines the single-objective Branin and Currin benchmarks. The
domain is \(\mathbf{x}=(x_1,x_2)\in[0,1]^2\). Let
$
\bar{x}_1 = 15x_1-5,~
\bar{x}_2 = 15x_2
$. 
We define the objective functions as
\[
f^1(\mathbf{x})
=
-
\left[
\left(
\bar{x}_2
-
\frac{5.1}{4\pi^2}\bar{x}_1^2
+
\frac{5}{\pi}\bar{x}_1
-
6
\right)^2
+
\left(10-\frac{10}{8\pi}\right)\cos(\bar{x}_1)
+
10
\right]
\]

\[
f^2(\mathbf{x})
=
-
\left[
\left(
1-\exp\left(-\frac{1}{2x_2}\right)
\right)
\frac{
2300x_1^3+1900x_1^2+2092x_1+60
}{
100x_1^3+500x_1^2+4x_1+20
}
\right]
\]
In implementation, \(x_2\) is clipped away from zero for numerical stability.

\paragraph{Welded Beam:} The unconstrained Welded Beam benchmark is a bi-objective structural design problem with four decision variables: weld thickness \(h\), weld length \(l\), beam height \(t\), and beam thickness \(b\). We use the unconstrained objective formulation  \cite{pymoo, COELLOCOELLO20021245}. The decision bounds are
\[
h\in[0.125,2],
\qquad
l\in[0.1,10],
\qquad
t\in[0.1,10],
\qquad
b\in[0.1,2]
\]
Writing \(\mathbf{x}=(h,l,t,b)\), the maximization objectives are
\[
f^1(\mathbf{x})
=
-
\left(
1.10471 h^2 l
+
0.04811 t b (14+l)
\right)
\]
\[
f^2(\mathbf{x})
=
-
\frac{2.1952}{t^3b}
\]

\paragraph{Vehicle Crash Safety:}
The Vehicle Crash Safety benchmark is a three-objective engineering design problem based on surrogate models for automobile crash worthiness \cite{Liao2008-cu,TANABE2020106078}. It has 5 design variables with bounds
$
x_i\in[1,3]$ for $i=1,\dots,5$.
The three maximization objectives are defined as
\[
f^1(\mathbf{x})
=
-
\left(
1640.2823
+2.3573285x_1
+2.3220035x_2
+4.5688768x_3
+7.7213635x_4
+4.4559504x_5
\right)
\]
\[
\begin{aligned}
f^2(\mathbf{x})
=
-\Big(
&6.5856075
-1.15x_1
-1.0427x_2
+0.9738x_3
+0.1365x_4
-1.05041x_1x_2  \\
&+0.06792x_1x_3
+0.54246x_1x_4
+1.0357x_2x_3
+0.8456x_2x_4
+1.4127x_3x_4
\Big)
\end{aligned}
\]
\[
f^3(\mathbf{x})
=
-
\left(
0.21886928
+0.00569696x_1x_4
+0.01086522x_2x_3
+0.0128098x_3x_4
\right)
\]

\paragraph{HPO (PD1 CIFAR-100):}
The hyperparameter optimization (HPO) benchmark is derived from the PD1 tabular surrogate suite \cite{wang2024hyperbo} and accessed through the \texttt{mfpbench} (\texttt{mf-prior-bench}) benchmark library (\url{https://github.com/automl/mf-prior-bench}). Specifically, we use the \texttt{PD1cifar100\_wideresnet\_2048} task, which models the multi-objective hyperparameter tuning of a Wide ResNet image classifier trained on CIFAR-100 with a batch size of 2048. Optimization is performed over the four-dimensional unit hypercube \(\mathbf{x} = (x_1, x_2, x_3, x_4) \in [0,1]^4\) (\(d=4\)), where \(\mathbf{x}\) maps to four learning-rate schedule and optimizer hyperparameters \(\boldsymbol{\theta}(\mathbf{x}) = (\theta_1, \theta_2, \theta_3, \theta_4)\) via linear and log-linear transformations:
\[
\begin{aligned}
\theta_1 &= \texttt{lr\_decay\_factor} = 0.010093 + x_1(0.989012 - 0.010093), \\
\theta_2 &= \texttt{lr\_initial} = \exp\!\big(\log(0.000010) + x_2\big(\log(9.779176) - \log(0.000010)\big)\big), \\
\theta_3 &= \texttt{lr\_power} = 0.100708 + x_3(1.999376 - 0.100708), \\
\theta_4 &= \texttt{opt\_momentum} = \exp\!\big(\log(0.000059) + x_4\big(\log(0.998993) - \log(0.000059)\big)\big).
\end{aligned}
\]
Each configuration \(\boldsymbol{\theta}(\mathbf{x})\) is queried at the full training budget of 199 epochs (\texttt{at=199}), returning the validation error \(\mathcal{E}_{\mathrm{val}}(\boldsymbol{\theta}(\mathbf{x}))\) and training cost \(\mathcal{C}_{\mathrm{train}}(\boldsymbol{\theta}(\mathbf{x}))\). Both quantities are negated to follow our maximization convention (\(M=2\)):
\[
f^1(\mathbf{x}) = -\mathcal{E}_{\mathrm{val}}\big(\boldsymbol{\theta}(\mathbf{x})\big),
\qquad
f^2(\mathbf{x}) = -\mathcal{C}_{\mathrm{train}}\big(\boldsymbol{\theta}(\mathbf{x})\big).
\]
\paragraph{Airfoil Design (AirDbM):}
The AirDbM benchmark represents an unconstrained two-objective aerodynamic shape optimization problem based on the Design-by-Morphing (DbM) parameterization of airfoil geometries \cite{10.1093/jcde/qwaf124,airfoils}. The benchmark is accessed from the open-source \texttt{AirDbM-Bench} repository (\url{https://github.com/jun9303/AirDbM-Bench/tree/main}). Optimization is conducted over a twelve-dimensional normalized decision vector \(\mathbf{x} \in [0,1]^{12}\) (\(d=12\)), which is linearly rescaled to the physical DbM morphing weight domain \(\tilde{\mathbf{x}} \in [-1, 1]^{12}\) across 12 baseline airfoils selected from the UIUC database via
\[
\tilde{x}_i = -1 + 2x_i,
\qquad
i = 1, \dots, 12.
\]
The morphing weights \(\tilde{\mathbf{x}}\) are normalized internally using an \(L_1\)-style absolute-sum scheme (\(\sum_{i=1}^{12} |\tilde{x}_i| = 1\), specified via \texttt{dbm\_normalization=\textquotesingle ABS\_SUM\textquotesingle}) to synthesize the morphed 2D airfoil geometry. Each candidate airfoil geometry is evaluated across an angle-of-attack sweep using the XFOIL aerodynamic solver, which accounts for the extended wall-clock runtimes observed across all methods on this benchmark in Table~1. The simulator directly returns two competing aerodynamic objectives (\(M=2\)) formulated for maximization:
\[
f^1(\mathbf{x}) = \left(\frac{C_l}{C_d}\right)_{\max}\!(\tilde{\mathbf{x}}),
\qquad
f^2(\mathbf{x}) = \Delta \theta_{\mathrm{stall}}(\tilde{\mathbf{x}}),
\]
where \((C_l / C_d)_{\max}\) denotes the maximum lift-to-drag ratio and \(\Delta \theta_{\mathrm{stall}}\) denotes the stall tolerance (stall margin), defined as the angle-of-attack range between the maximum lift-to-drag operating point and the stall angle (the first local maximum of the lift coefficient \(C_l\)).

\paragraph{Penicillin Production:}
The Penicillin production benchmark is a three-objective biochemical process optimization problem originally formulated by Liang and Lai \cite{liang2021penicillin}. We utilize the BoTorch implementation \cite{balandat2020botorch} (\texttt{botorch.test\_functions.multi\_objective.Penicillin}). The optimization is conducted over a seven-dimensional normalized decision vector \(\mathbf{x} \in [0,1]^7\) (\(d=7\)), which is linearly mapped to the native engineering bounds \([\mathbf{l}, \mathbf{u}] \subset \mathbb{R}^7\) of the simulator via
\[
\tilde{x}_i = l_i + x_i(u_i - l_i),
\qquad
i = 1, \dots, 7,
\]
corresponding to seven fermentation control variables (culture volume, biomass concentration, temperature, glucose substrate concentration, substrate feed rate, substrate feed concentration, and \(\mathrm{H}^+\) concentration).
The optimization problem returns the three-objective (\(M=3\)) maximization vector \(\mathbf{f}(\mathbf{x}) = \big(Y_{\mathrm{pen}}(\tilde{\mathbf{x}}),\, -E_{\mathrm{CO}_2}(\tilde{\mathbf{x}}),\, -T_{\mathrm{ferm}}(\tilde{\mathbf{x}})\big)\), where \(Y_{\mathrm{pen}}\) denotes penicillin yield, \(E_{\mathrm{CO}_2}\) denotes \(\mathrm{CO}_2\) production, and \(T_{\mathrm{ferm}}\) denotes fermentation time.

\begin{table}[h]
\centering
\caption{Input dimensions and number of objectives for each benchmark problem. Values in parentheses denote additional configurations evaluated in the scaling studies in Appendix~\ref{app:scaling}.}
\label{tab:test-functions}
\begin{tabular}{lcc}
\toprule
Benchmark Problem & Dimensions ($d$) & No.\ of objectives ($M$) \\
\midrule
ZDT2 & 9 & 2 \\
ZDT4 & 6 & 2 \\
ZDT6 & 6 & 2 \\
DTLZ1 & 8 (10, 15, 20) & 3 (2, 3, 4) \\
DTLZ2 & 6 & 3 \\
DTLZ3 & 4 (10, 15, 20) & 2 (2, 3, 4) \\
DTLZ7 & (6, 10, 15, 20) & (2, 3, 4) \\
Branin--Currin & 2 & 2 \\
Welded Beam & 4 & 2 \\
Vehicle Crash & 5 & 3 \\
HPO (PD1 CIFAR-100) & 4 & 2 \\
Airfoil Design (AirDbM) & 12 & 2 \\
Penicillin Production & 7 & 3 \\
\bottomrule
\end{tabular}
\end{table}


\newpage
\section{Additional Results And Scaling Studies}
\label{app:results}
This appendix presents additional empirical results and studies examining performance as the input dimension and number of objectives increase, further supporting the findings in the main paper.

\subsection{Batch Mode ($Q=3$) IGD Results}

\begin{figure}[!hbt]
\includegraphics[width=0.985\linewidth]{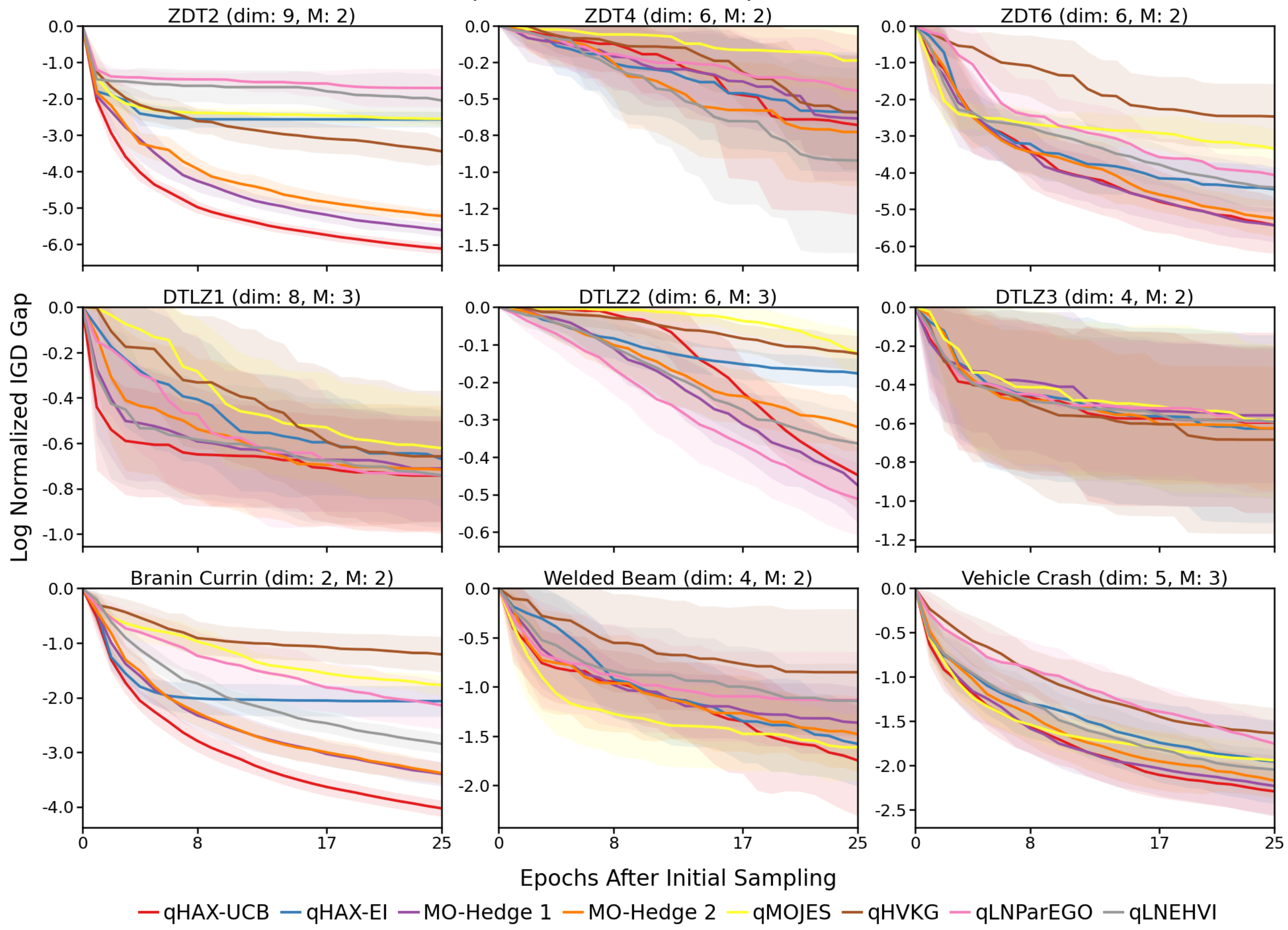}
\centering
\vspace{-2mm}
\caption{Log-normalized inverted generational distance gap for batch optimization (\(Q=3\)) across benchmarks. Lower values are better.
Mean (solid line)  \(\pm 1\) standard deviation (shaded)  over 20 runs.}
\label{fig:q3igd}
\vspace{-3mm}
\end{figure}

Figure~\ref{fig:q3igd} reports the log‑normalized inverted generational distance (IGD) results for all benchmark problems in the batch setting (\(Q=3\)). IGD complements hypervolume by directly measuring the geometric distance between the current Pareto approximation and a reference Pareto front (typically the true Pareto front or a high‑fidelity approximation), thereby emphasizing both convergence and coverage. Lower values correspond to more accurate and more uniformly covered Pareto‑front approximations.

Across the benchmarks, the IGD results exhibit trends similar to those observed for hypervolume. \textit{q}HAX‑$UCB$ consistently achieves strong performance with relatively low variance, indicating rapid convergence and robust coverage of the Pareto front. \textit{q}HAX‑$EI$ also exhibits competitive behavior, though with greater problem dependence. In contrast, the baseline state-of-the-art MOBO approaches show more variable behavior across benchmarks. As in the hypervolume results, no single baseline method dominates uniformly, reinforcing the sensitivity of acquisition performance to problem structure. The hedge‑based strategies maintain competitive and stable IGD performance across all benchmarks and avoid the stagnation observed in some baseline methods.

Overall, these IGD results for batch optimization are consistent with the main findings of the paper. The \textit{q}HAX‑$\alpha$ framework achieves robust Pareto‑front coverage in addition to strong hypervolume performance, while MO‑Hedge further stabilizes optimization by adapting acquisition selection online. Importantly, the results confirm that no single acquisition strategy attains uniformly optimal IGD across all benchmarks, whereas hedge‑based selection mitigates this variability and yields consistently competitive Pareto approximations.

\subsection{Sequential Mode ($Q=1$) Benchmarking Results}

\begin{figure}[!hbt]
\includegraphics[width=0.985\linewidth]{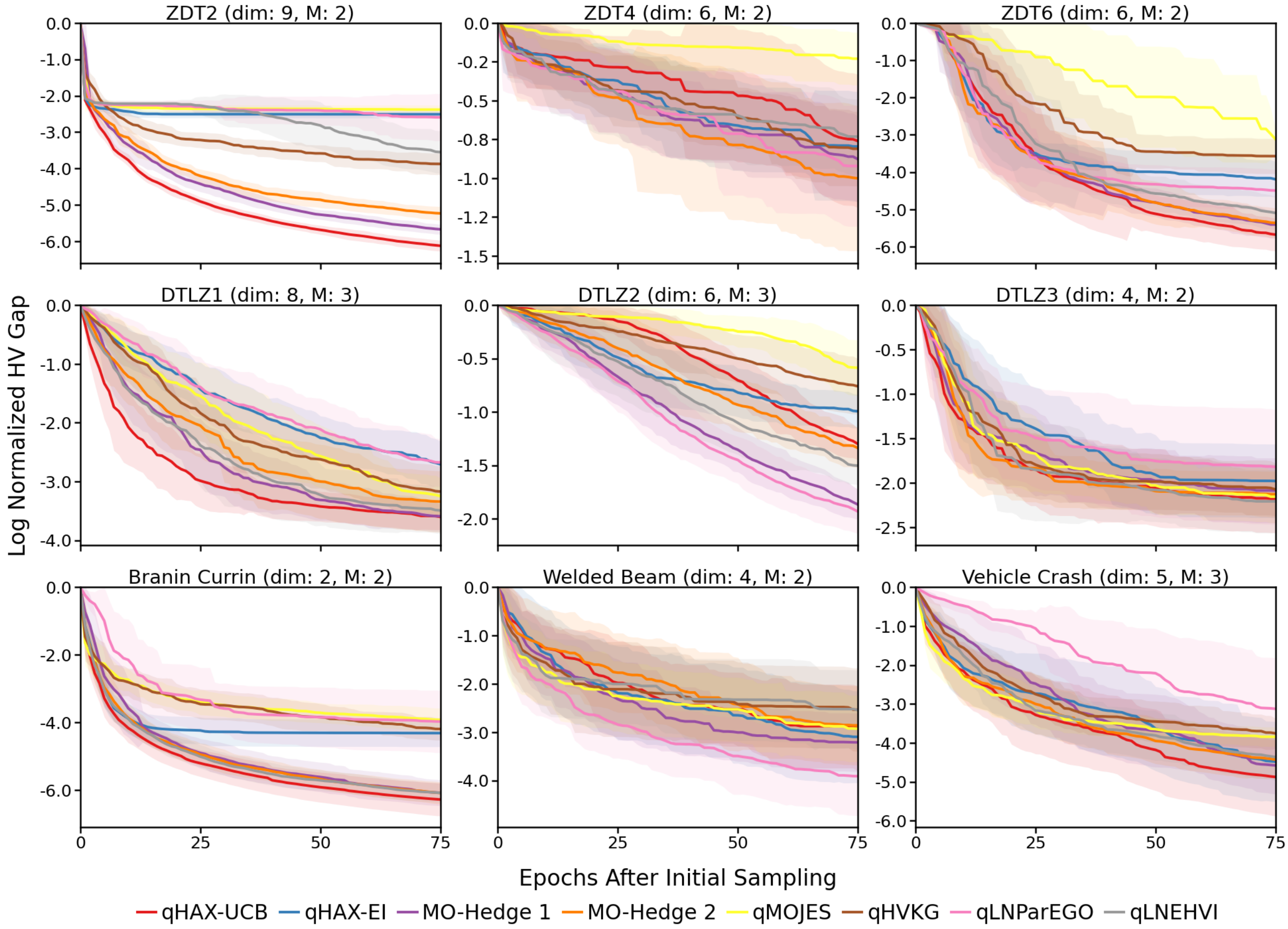}
\centering
\vspace{-2mm}
\caption{Log-normalized hypervolume gap for sequential optimization (\(Q=1\)) across benchmarks. Lower values are better.
Mean (solid line)  \(\pm 1\) standard deviation (shaded)  over 20 runs.}
\label{fig:q1hv}
\vspace{-3mm}
\end{figure}

\begin{figure}[!hbt]
\vspace{4mm}
\includegraphics[width=0.985\linewidth]{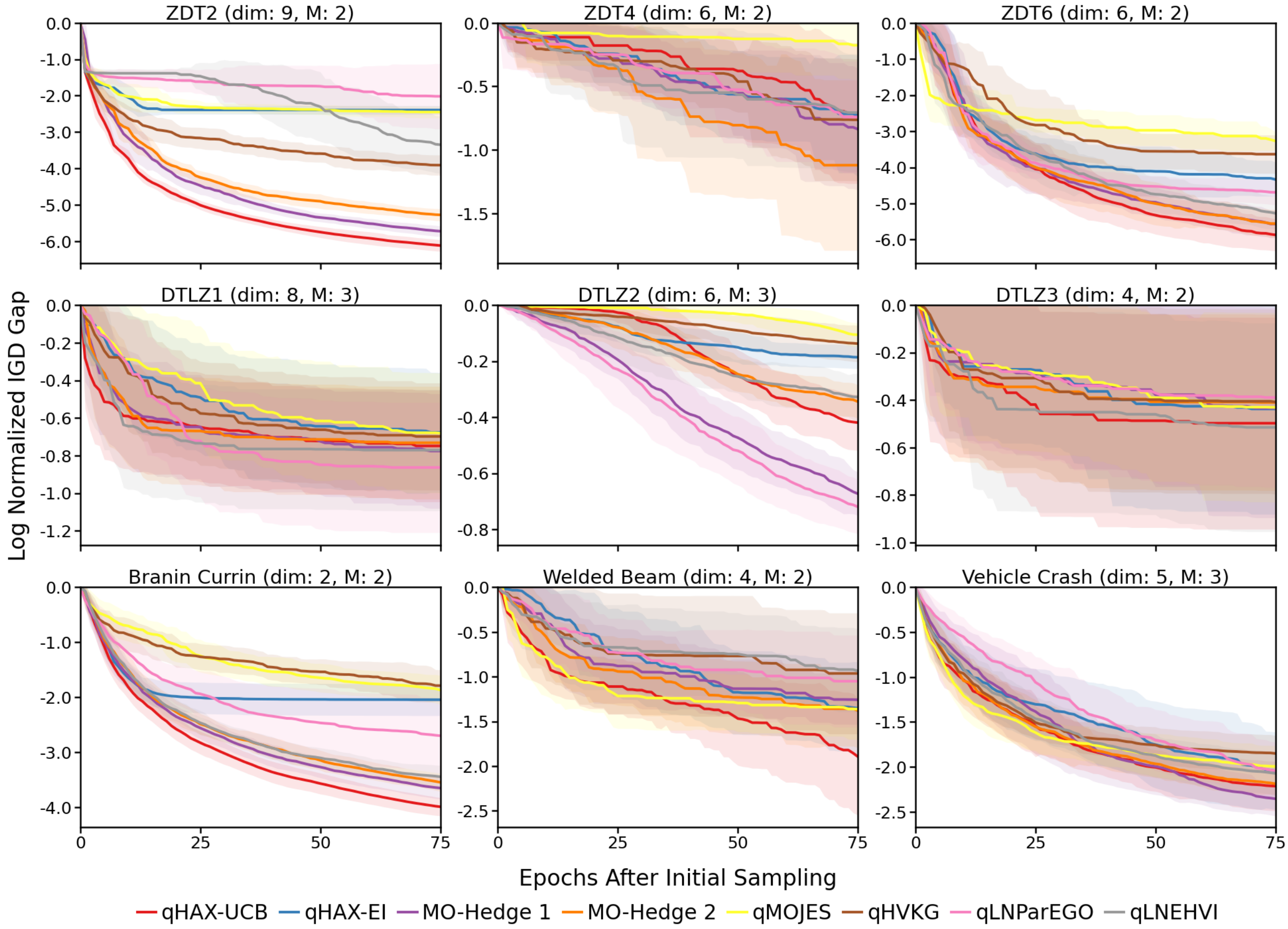}
\centering
\vspace{-2mm}
\caption{Log-normalized IGD gap for sequential optimization (\(Q=1\)) across benchmarks. Lower values are better.
Mean (solid line)  \(\pm 1\) standard deviation (shaded)  over 20 runs.}
\label{fig:q1igd}
\vspace{-3mm}
\end{figure}

In addition to evaluating acquisition strategies in parallel, we also benchmark them in the sequential BO setting (\(Q=1\)). All experimental details are identical to those used in the batch setting, except that evaluations are performed sequentially, yielding a total of 75 optimization epochs. As in the batch experiments, we report the log-normalized hypervolume (Fig.~\ref{fig:q1hv}) and log-normalized IGD (Fig.~\ref{fig:q1igd}) trajectories across all benchmark problems. By removing the complexities associated with batch candidate selection, the sequential setting provides a clearer view of each acquisition strategy’s intrinsic exploration--exploitation behavior.

Across the benchmarks, the sequential results exhibit trends consistent with those observed in batch mode. On most problems, \textit{q}HAX-$UCB$ achieves the fastest reduction in both hypervolume and IGD gap while maintaining low variance, indicating efficient convergence and robust Pareto-front coverage. \textit{q}HAX-$EI$ remains competitive but converges more slowly on more challenging problems, particularly where sustained exploration might be required. Baseline methods show mixed behavior: scalarization-based and noisy hypervolume approaches perform reasonably on simpler problems but display slower convergence or higher variance on benchmarks with multimodal or non-uniform Pareto structure. Entropy-based (\textit{q}MOJES) and lookahead-based (\textit{q}HVKG) methods generally exhibit slower improvement and greater variability, reflecting their sensitivity to model uncertainty and objective interactions in the sequential setting. As in batch mode, no baseline acquisition dominates across all benchmark tasks.

Both variants of MO-Hedge perform consistently well across the benchmarking suite. Although no single acquisition strategy is uniformly optimal, MO-Hedge is able to match or track the performance of its strongest constituent on each problem. This further supports our claim that, for black-box objectives with unknown structure, hedging over a portfolio of acquisition strategies is an effective mechanism for achieving robust and efficient performance.

Overall, the sequential results corroborate the main findings of the paper. The \textit{q}HAX-$\alpha$ framework delivers strong and stable performance without relying on bespoke multi-objective acquisition functions, while MO-Hedge further enhances robustness by adapting acquisition selection online. Importantly, the consistency between sequential (\(Q=1\)) and batch (\(Q=3\)) results indicates that the benefits of Pareto-aware selection and hedge-based adaptation are not artifacts of batch optimization but reflect general advantages across multi-objective Bayesian optimization settings.

\paragraph{Runtime Analysis:}

Table~\ref{tab:runtime_comparison} reports the average wall-clock runtime for each acquisition strategy in the sequential setting. Overall, \textit{q}HAX‑$\alpha$ exhibits stable and predictable runtime across all benchmark problems, with both \textit{q}HAX‑$UCB$ and \textit{q}HAX‑$EI$ showing only modest variation across problem dimensionality and number of objectives. In contrast, more specialized MOBO baselines exhibit substantially higher and more variable computational costs.
\begin{wraptable}{r}{0.66\textwidth}
\centering
\scriptsize
\setlength{\tabcolsep}{2pt}
\begin{tabular}{lcccccc}
\toprule
Test Function & \textit{q}HAX-$UCB$ & \textit{q}HAX-$EI$ & \textit{q}MOJES & \textit{q}HVKG & \textit{q}LNParEGO & \textit{q}LNEHVI \\
\midrule
ZDT2 & $676.62$ & $715.49$ & $2945.64$ & $558.25$ & $126.52$ & $185.74$ \\
ZDT4 & $797.93$ & $613.77$ & $4469.76$ & $668.50$ & $265.70$ & $186.82$ \\
ZDT6 & $597.61$ & $612.06$ & $17750.69$ & $817.01$ & $228.30$ & $523.60$ \\
DTLZ1 & $685.13$ & $717.91$ & $6884.48$ & $1074.03$ & $378.68$ & $453.49$ \\
DTLZ2 & $668.83$ & $684.98$ & $6523.09$ & $959.74$ & $270.95$ & $574.83$ \\
DTLZ3 & $667.09$ & $647.52$ & $4010.11$ & $615.54$ & $167.90$ & $127.60$ \\
Branin--Currin & $721.12$ & $729.20$ & $3190.28$ & $420.43$ & $148.63$ & $176.83$ \\
Welded Beam & $576.17$ & $584.69$ & $4946.90$ & $708.45$ & $232.74$ & $138.66$ \\
Vehicle Crash & $775.76$ & $674.66$ & $5859.59$ & $792.02$ & $352.50$ & $510.71$ \\
\bottomrule
\end{tabular}
\caption{Mean runtime in seconds for each algorithm and test function over 20 independent runs. (Sequential mode: $Q=1$)}
\label{tab:runtime_comparison}
\end{wraptable}
In particular, the entropy‑based \textit{q}MOJES baseline is consistently the most expensive method, often exceeding the runtime of \textit{q}HAX‑$\alpha$ by an order of magnitude, reflecting the cost of Pareto‑set sampling and entropy estimation. Lookahead‑based methods such as \textit{q}HVKG also incur elevated runtime due to their one‑step Bayes‑optimal optimization. Scalarization‑based and noisy hypervolume‑based methods are comparatively efficient in the sequential setting, benefiting from the absence of batch coupling and exhibiting lower overall runtime.

Importantly, the runtime of \textit{q}HAX‑$\alpha$ remains largely insensitive to problem structure, number of objectives, or batch size, as its dominant cost arises from the inner Pareto‑front generation. These results broadly mirror the batch‑mode runtime analysis and highlight \textit{q}HAX‑$\alpha$ as an efficient and scalable alternative for sequential multi‑objective Bayesian optimization.

\subsection{Input-Dimension and Objective-Count Scaling}
\label{app:scaling}

\begin{figure}[!b]
    \centering
    \vspace{-2mm}
    \includegraphics[width=0.985\textwidth]{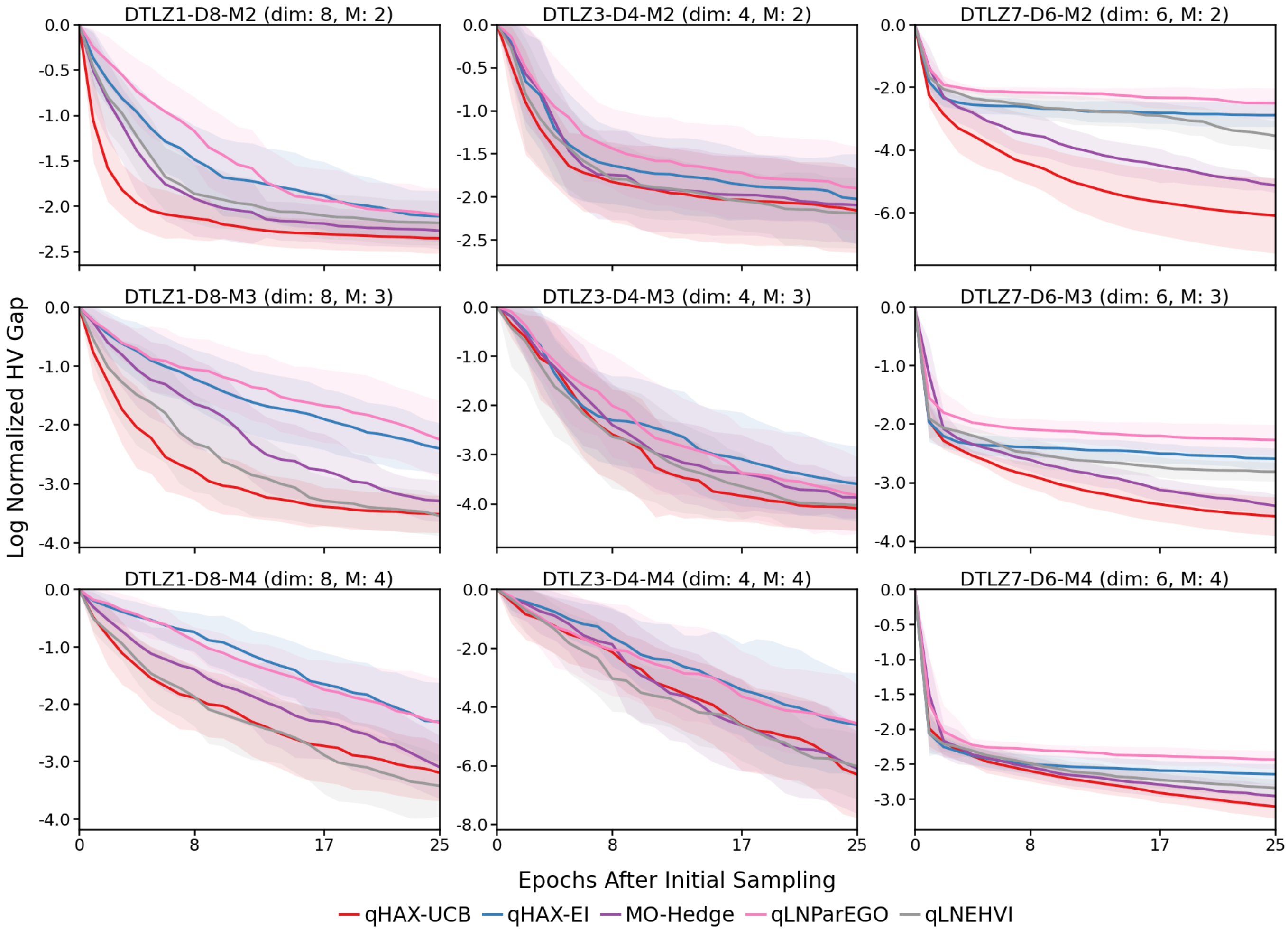}
    \caption{Log-normalized hypervolume gap for objective-count scaling in the batch setting ($Q=3$). Rows correspond to $M\in\{2,3,4\}$ objectives for DTLZ1, DTLZ3, and DTLZ7. Lower values are better. Mean (solid line)  \(\pm 1\) standard deviation (shaded)  over 20 runs.}
    \label{fig:objective_scaling}
    \vspace{-3mm}
\end{figure}

We further examine the performance of the acquisition strategies as the input dimension and number of objectives increase. All scaling experiments use the parallel batch setting ($Q=3$), with the same initial design, evaluation budget, and noise model as the main experiments. We consider DTLZ1, DTLZ3, and DTLZ7, which provide complementary optimization challenges. In particular, DTLZ1 and DTLZ3 contain multiple local Pareto fronts, while DTLZ7 has a disconnected Pareto front and is included to evaluate performance on a geometry not represented in the main benchmark suite. The tested ranges, $d\in\{10,15,20\}$ and $M\in\{2,3,4\}$, provide representative increases in input dimension and objective count under the fixed evaluation budget. Performance is reported using the log-normalized hypervolume gap, with lower values indicating better performance. To keep the scaling study computationally tractable, we do not include \textit{q}MOJES and \textit{q}HVKG, which incurred high or variable computational costs without providing a consistent performance advantage in the main benchmark suite. The study therefore retains \textit{q}LNParEGO and \textit{q}LNEHVI as representative specialized MOBO baselines.

Figure~\ref{fig:objective_scaling} examines objective-count scaling for $M\in\{2,3,4\}$ while retaining the baseline input dimension for each problem. Across these settings, \textit{q}HAX-$UCB$ maintains strong and generally consistent performance, achieving the lowest or near-lowest final hypervolume gap on most configurations. The advantage is particularly clear on DTLZ7, where \textit{q}HAX-$UCB$ achieves a substantially lower final gap than the other methods for all tested values of $M$. On DTLZ1 and DTLZ3, \textit{q}HAX-$UCB$ remains among the strongest methods as the number of objectives increases. \textit{q}LNEHVI remains competitive on both problems but performs weakly on DTLZ7. \textit{q}HAX-$EI$ is more problem dependent and generally converges more slowly, consistent with the behavior observed in the main benchmark suite. MO-Hedge remains competitive across the tested objective counts and typically tracks the stronger acquisition strategies without substantial degradation as $M$ increases.

\begin{figure}[t]
    \centering
    \vspace{-2mm}
    \includegraphics[width=0.985\textwidth]{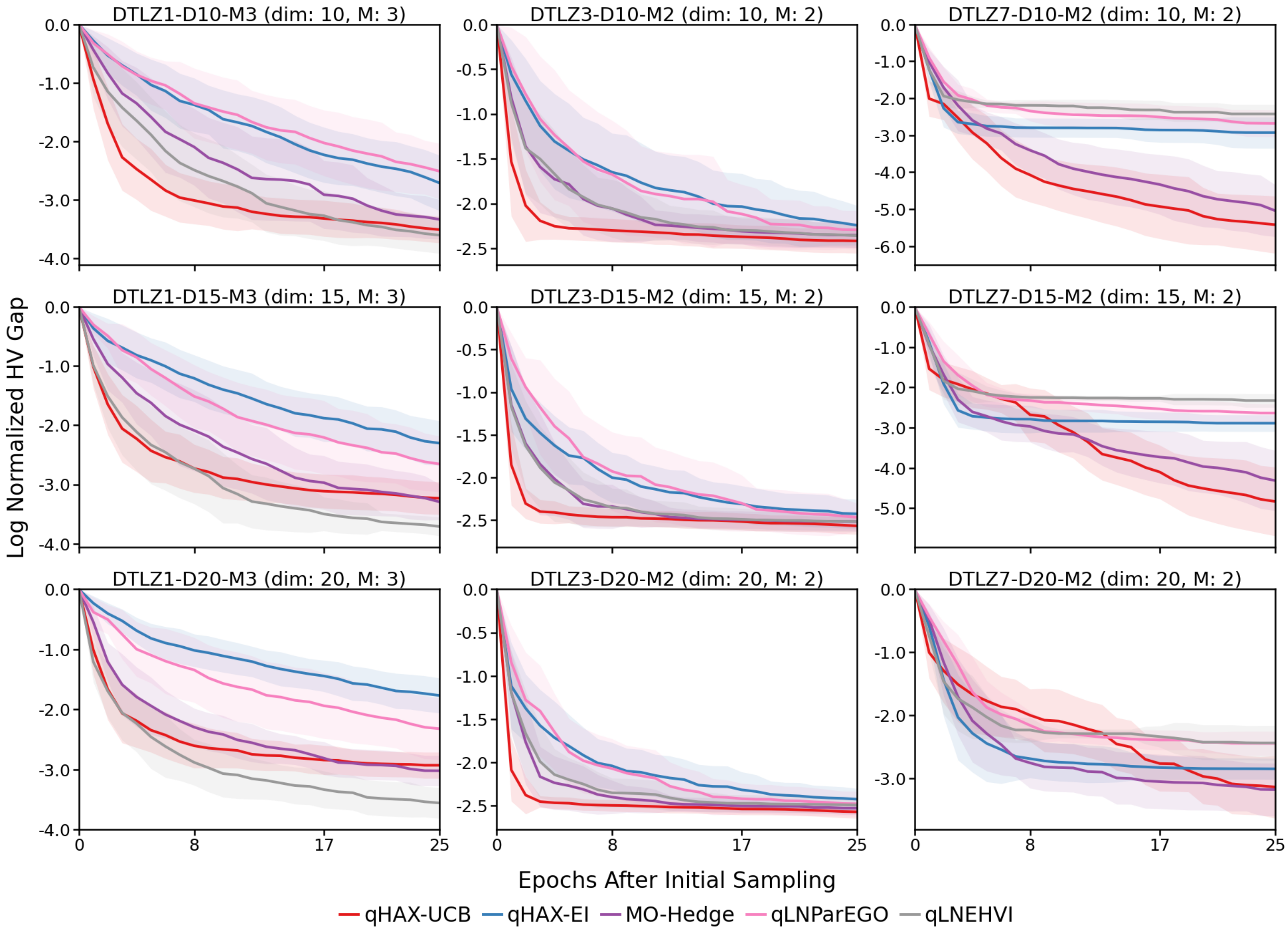}
    \caption{Log-normalized hypervolume gap for input-dimension scaling in the batch setting ($Q=3$). Rows correspond to $d\in\{10,15,20\}$ for DTLZ1, DTLZ3, and DTLZ7. Lower values are better. Mean (solid line)  \(\pm 1\) standard deviation (shaded)  over 20 runs.}
    \label{fig:dimension_scaling}
    \vspace{-3mm}
\end{figure}

Figure~\ref{fig:dimension_scaling} examines input-dimension scaling for $d\in\{10,15,20\}$. The strong performance of \textit{q}HAX-$UCB$ is maintained as the input dimension increases, with rapid initial improvement and the lowest or near-lowest final hypervolume gap across most DTLZ3 and DTLZ7 configurations. On DTLZ1, \textit{q}LNEHVI becomes particularly competitive at $d=15$ and $d=20$, while \textit{q}HAX-$UCB$ remains among the strongest methods. MO-Hedge also remains competitive across dimensions, while \textit{q}HAX-$EI$ and \textit{q}LNParEGO show greater problem dependence.

We note that the NSGA-II population size and number of generations were held fixed at their baseline values across all scaling experiments to ensure a consistent comparison. Increasing this internal optimization budget could produce a more accurate acquisition-space Pareto approximation and potentially improve \textit{q}HAX-$\alpha$ performance in the higher-dimensional and higher-objective settings. Overall, \textit{q}HAX-$\alpha$ and MO-Hedge maintain strong performance across the tested increases in input dimension and number of objectives, with particularly strong results on DTLZ7 and its disconnected Pareto front. These results support their empirical scalability within the tested regime of $d\leq20$ and $M\leq4$. The computational scalability of \textit{q}HAX-$\alpha$ is further demonstrated by the runtime analysis.

\paragraph{Runtime Analysis:}

\begin{table*}[b]
\centering
\scriptsize
\setlength{\tabcolsep}{2pt}
\resizebox{\textwidth}{!}{%
\begin{tabular}{lccrrrr@{\hspace{2mm}}|@{\hspace{2mm}}lccrrrr}
\toprule
\multicolumn{7}{c}{\textit{Objective-count scaling}} &
\multicolumn{7}{c}{\textit{Input-dimension scaling}} \\
\cmidrule(lr){1-7}\cmidrule(lr){8-14}
Problem & $d$ & $M$ & \textit{q}HAX-$UCB$ & \textit{q}HAX-$EI$ & \textit{q}LNParEGO & \textit{q}LNEHVI &
Problem & $d$ & $M$ & \textit{q}HAX-$UCB$ & \textit{q}HAX-$EI$ & \textit{q}LNParEGO & \textit{q}LNEHVI \\
\midrule
DTLZ1 & 8 & 2 & $282.69$ & $273.38$ & $552.52$ & $687.39$ &
DTLZ1 & 10 & 3 & $370.81$ & $349.51$ & $1134.73$ & $1391.13$ \\
DTLZ3 & 4 & 2 & $251.50$ & $252.81$ & $561.85$ & $670.51$ &
DTLZ3 & 10 & 2 & $306.40$ & $290.71$ & $1130.09$ & $1074.24$ \\
DTLZ7 & 6 & 2 & $257.76$ & $264.18$ & $426.40$ & $647.62$ &
DTLZ7 & 10 & 2 & $232.75$ & $236.45$ & $976.51$ & $1265.02$ \\
DTLZ1 & 8 & 3 & $293.43$ & $291.60$ & $529.33$ & $1128.58$ &
DTLZ1 & 15 & 3 & $439.34$ & $417.71$ & $1114.13$ & $1542.25$ \\
DTLZ3 & 4 & 3 & $266.28$ & $269.83$ & $532.34$ & $945.88$ &
DTLZ3 & 15 & 2 & $378.91$ & $371.69$ & $1190.14$ & $1304.93$ \\
DTLZ7 & 6 & 3 & $265.29$ & $263.69$ & $413.16$ & $1163.43$ &
DTLZ7 & 15 & 2 & $240.90$ & $250.71$ & $1049.41$ & $1539.39$ \\
DTLZ1 & 8 & 4 & $291.28$ & $287.48$ & $425.62$ & $2560.08$ &
DTLZ1 & 20 & 3 & $492.56$ & $465.32$ & $1099.30$ & $1790.61$ \\
DTLZ3 & 4 & 4 & $262.74$ & $262.91$ & $449.64$ & $2262.56$ &
DTLZ3 & 20 & 2 & $417.37$ & $411.77$ & $1136.85$ & $1491.83$ \\
DTLZ7 & 6 & 4 & $263.48$ & $259.37$ & $313.80$ & $2491.70$ &
DTLZ7 & 20 & 2 & $253.18$ & $256.31$ & $1069.42$ & $1660.36$ \\
\bottomrule
\end{tabular}%
}
\caption{Mean runtime in seconds for objective-count and input-dimension scaling over 20 independent runs (batch mode: $Q=3$).}
\label{tab:scaling_runtime_comparison}
\end{table*}

Table~\ref{tab:scaling_runtime_comparison} reports the average wall-clock time required for each method across the input-dimension and objective-count scaling studies. The \textit{q}HAX-$\alpha$ variants are the most computationally efficient methods across all tested configurations. Their runtimes remain nearly constant as the number of objectives increases from $M=2$ to $M=4$, indicating that the additional objective-wise acquisition evaluations introduce limited overhead relative to the fixed NSGA-II candidate-generation budget. Runtime increases more noticeably with input dimension, but remains below 500 seconds at $d=20$ and substantially lower than the included specialized MOBO baselines. \textit{q}LNParEGO also exhibits stable runtime across the tested configurations, whereas \textit{q}LNEHVI becomes increasingly expensive as the number of objectives and input dimensions increase. Overall, \textit{q}HAX-$\alpha$ exhibits the strongest computational scaling across the tested regime, remaining the fastest approach in every configuration while showing only modest runtime growth with increasing input dimension and negligible growth with objective count.

\newpage
\section{Ablation Studies And Further Discussion}
\label{app:ablations}

This appendix provides targeted ablation studies and behavioral diagnostics for both \textit{q}HAX-$\alpha$ and MO-Hedge. Appendix~\ref{app:acqablations} isolates the respective contributions of acquisition-space Pareto candidate generation and posterior-mean hypervolume ranking in \textit{q}HAX-$\alpha$, and visualizes the resulting search dynamics against baseline MOBO strategies on the Branin--Currin benchmark. Appendix~\ref{app:hedgeablations} evaluates alternative historical reward-assignment rules for MO-Hedge and examines the evolution of portfolio selection probabilities over the course of optimization.

\subsection{\textit{q}HAX-$\alpha$ Ablations and Analysis}
\label{app:acqablations}

\paragraph{Component and Candidate-Ranking Ablations:}
The \textit{q}HAX-$\alpha$ framework decomposes multi-objective acquisition into two sequential stages (Algorithm~1): (i) generating a Pareto-optimal candidate set $\mathcal{C}_t^0$ in acquisition space by optimizing the vector-valued acquisition $\boldsymbol{\alpha}_t(\mathbf{x})$, and (ii) greedily selecting a batch of $Q$ candidates from $\mathcal{C}_t^0$ according to their posterior-mean hypervolume contribution $h_t^q(\mathbf{c}_t)$. To quantify the individual contribution of each stage and assess whether posterior-mean hypervolume ranking can be replaced by simpler diversity or scalar aggregation heuristics, we compare the complete \textit{q}HAX-$\alpha$ algorithm (\texttt{Full}) against five ablated variants for both \textit{q}HAX-UCB and \textit{q}HAX-EI across all nine synthetic benchmarks ($Q=3$, 20 paired runs):
\begin{enumerate}
    \item \textit{Matched Random Pool with HV Ranking} (\texttt{Rand-HV}): Runs NSGA-II on $\boldsymbol{\alpha}_t(\mathbf{x})$ to determine the size $|\mathcal{C}_t^0|$ of the nondominated acquisition set, but replaces $\mathcal{C}_t^0$ with an equal number of i.i.d.\ uniform random points in $\mathcal{X}$ while retaining greedy posterior-mean hypervolume selection. This variant tests whether posterior-mean hypervolume ranking alone is sufficient without acquisition-space Pareto candidate generation.
    \item \textit{Acquisition-Pareto Pool with Random Selection}(\texttt{Pareto-Rand}): Retains the NSGA-II acquisition-Pareto candidate set $\mathcal{C}_t^0$, but selects $Q$ candidates uniformly at random without replacement, removing the hypervolume ranking stage.
    \item \textit{Objective-Space Distance Ranking} (\texttt{Obj-Dist}): Retains $\mathcal{C}_t^0$, but replaces hypervolume ranking with greedy maximin Euclidean distance selection in standardized posterior-mean objective space relative to the evaluated set $\boldsymbol{\mu}_t(\mathcal{D}_t)$, updating the comparison set after each selection within the batch.
    \item \textit{Decision-Space Distance Ranking} (\texttt{Input-Dist}): Retains $\mathcal{C}_t^0$, but selects $Q$ candidates via greedy maximin Euclidean distance in normalized decision space $[0,1]^d$ relative to the evaluated inputs.
    \item \textit{Summed Acquisition Ranking} (\texttt{Acq-Sum}): Retains $\mathcal{C}_t^0$, but ranks candidates by the unweighted scalar sum of their per-objective acquisition values, $\sum_{m=1}^M \alpha_t^m(\mathbf{c}_t)$, selecting the top $Q$ candidates. This is closest to the MOBO strategy introduced in~\cite{mixmobo}.
\end{enumerate}

\begin{figure}[!hbt]
\includegraphics[width=0.985\linewidth]{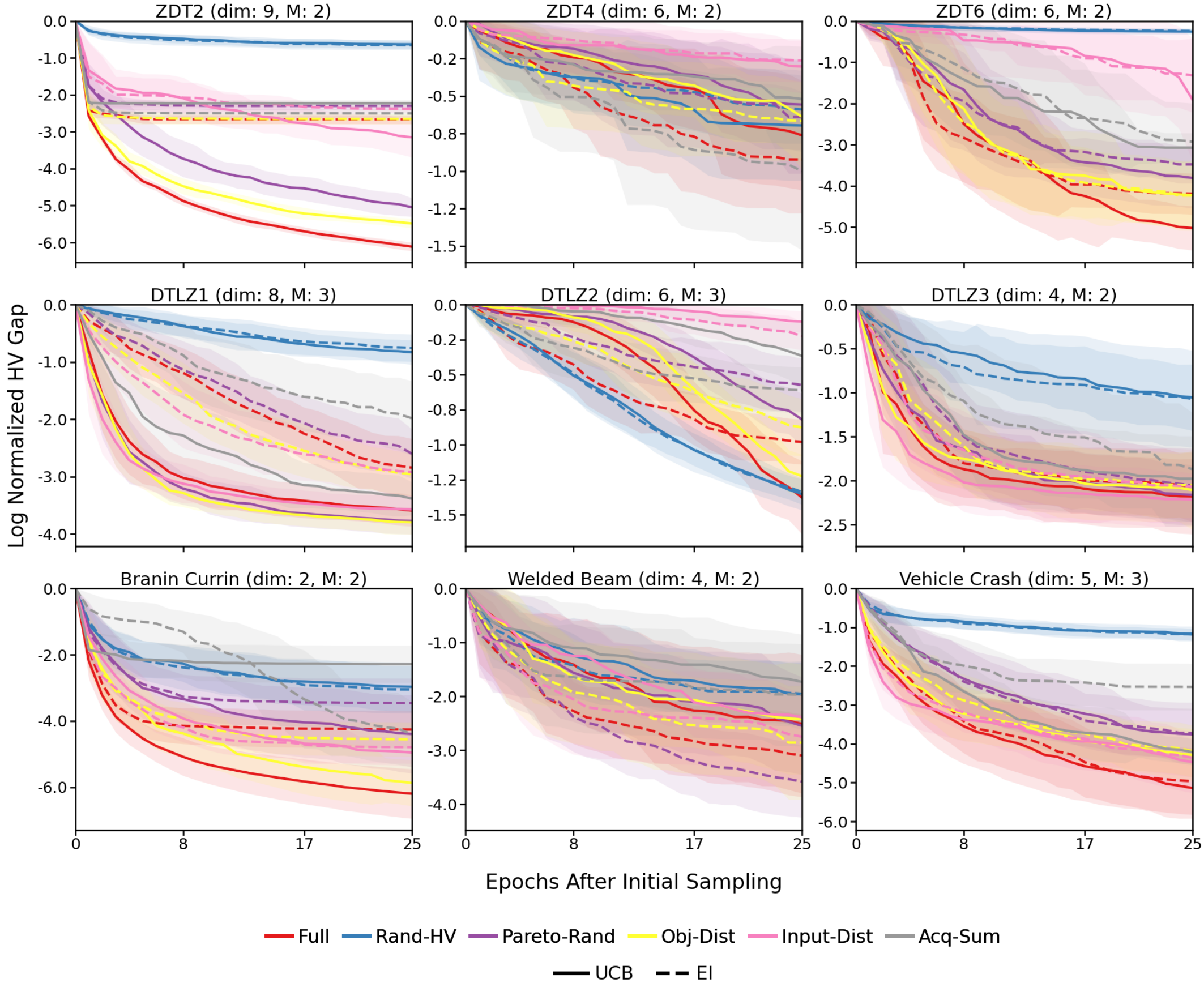}
\centering
\vspace{-2mm}
\caption{Log-normalized hypervolume gap for \textit{q}HAX-$\alpha$ variants using batch setting (\(Q=3\)). Lower values are better.
Mean (solid and dashed line) \(\pm 1\) standard deviation (shaded) over 20 runs.}
\label{fig:ablateaf}
\vspace{-3mm}
\end{figure}

Figure~\ref{fig:ablateaf} reports the log-normalized hypervolume gap trajectories for all twelve variants (solid lines for \textit{q}HAX-UCB and dashed lines for \textit{q}HAX-EI). The results reveal a clear hierarchy between the two stages of the framework. Replacing the acquisition-Pareto candidate pool with a matched-size random pool (\texttt{Rand-HV}) causes the largest and most consistent performance degradation across the benchmark suite: \texttt{Full} \textit{q}HAX-UCB outperforms \texttt{Rand-HV} on all nine benchmarks, and \texttt{Full} \textit{q}HAX-EI outperforms it on eight of the nine benchmarks. Because posterior-mean hypervolume ranking is purely exploitative when applied to uninformative candidates, it cannot recover the uncertainty-aware exploration provided by optimizing $\boldsymbol{\alpha}_t(\mathbf{x})$. This confirms that acquisition-space Pareto candidate generation is the primary driver of exploration and sample efficiency in \textit{q}HAX-$\alpha$.

Retaining the acquisition-Pareto candidate set $\mathcal{C}_t^0$ but selecting candidates at random (\texttt{Pareto-Rand}) yields substantially better performance than \texttt{Rand-HV}, yet still trails \texttt{Full} \textit{q}HAX-UCB on seven of nine benchmarks and \texttt{Full} \textit{q}HAX-EI on eight of nine benchmarks, most notably on ZDT2, ZDT6, DTLZ2, Branin--Currin, and Vehicle Crash. Among the alternative ranking rules applied to $\mathcal{C}_t^0$, greedy objective-space maximin distance (\texttt{Obj-Dist}) is the most competitive heuristic, tracking \texttt{Full} closely on several problems by promoting spread in predicted objective space; however, because it promotes pure geometric spread in objective space without accounting for Pareto dominance direction, it underperforms hypervolume ranking on problems such as ZDT2, ZDT4, ZDT6, Branin--Currin, and Vehicle Crash. Decision-space distance ranking (\texttt{Input-Dist}) and summed-acquisition ranking (\texttt{Acq-Sum}) degrade performance more severely, as uniform input-space spreading wastes evaluations in dominated regions while summing acquisition values collapses trade-offs toward extreme single-objective optima. Overall, these ablations demonstrate that both stages of \textit{q}HAX-$\alpha$ are essential and complementary: Pareto optimization of $\boldsymbol{\alpha}_t(\mathbf{x})$ critically generates an informative, uncertainty-aware candidate front, while posterior-mean hypervolume ranking provides a balanced rule for resolving objective trade-offs across the candidate pool.

\begin{figure}[t!]
    \centering
    \includegraphics[width=0.985\textwidth]{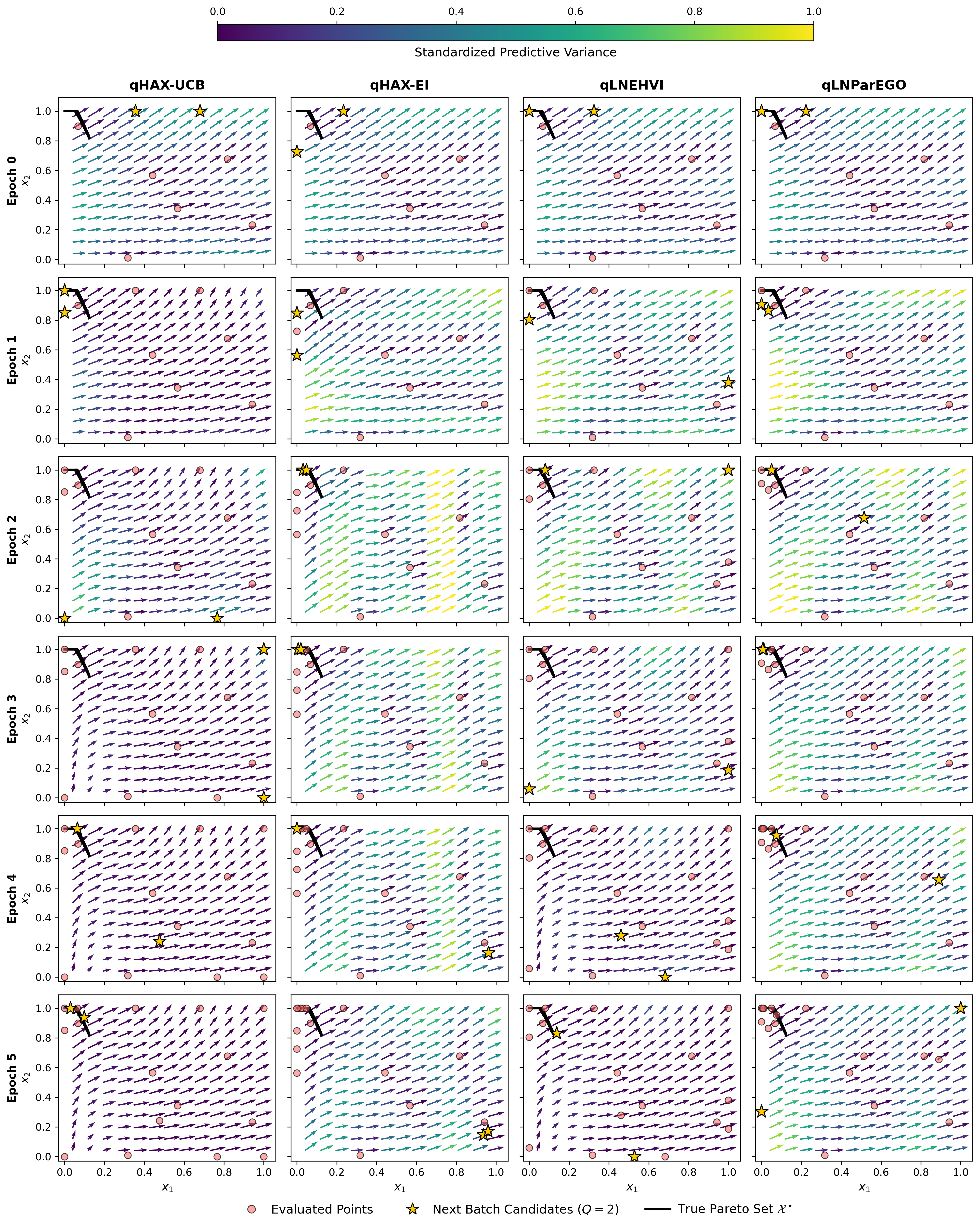}
    \caption{Evolution of GP posterior-mean objective vector fields (quiver arrows, colored by standardized predictive variance) and batch candidate selection ($Q=2$, gold stars) from $n_0=6$ initial points across Epochs 0--5 on Branin--Currin.}
    \label{fig:branin_quiver}
\end{figure}

\paragraph{Behavioral Analysis on Branin--Currin:}
To provide geometric intuition for the differing behaviors of \textit{q}HAX-$\alpha$ relative to standard scalarization and noisy expected hypervolume improvement baselines, Figure~\ref{fig:branin_quiver} visualizes the evolution of the fitted GP surrogates with $Q=2$ batch candidate selections for \textit{q}HAX-UCB, \textit{q}HAX-EI, \textit{q}LNEHVI, and \textit{q}LNParEGO on the two-dimensional Branin--Currin benchmark ($d=2, M=2$). All four methods are initialized from the same $n_0 = 6$ Halton design points (red circles) at Epoch~0 and tracked across Epochs~0--5. In each panel, the quiver arrows depict the GP posterior-mean objective vector $(\mu_t^1(\mathbf{x}), \mu_t^2(\mathbf{x}))$ (rescaled to $[0,1]^2$ so that longer arrows pointing toward the top-right indicate superior multi-objective performance), while the arrow color encodes the mean standardized predictive variance across the two objectives. The true Pareto set $\mathcal{X}^\star$, which occupies a compact curve in the upper-left corner, is shown as a solid black line, and the $Q=2$ candidates selected at each epoch are marked by gold stars.

In analyzing Figure~\ref{fig:branin_quiver}, we refrain from making strong quantitative claims about the behavior of each acquisition strategy, since the visualization depicts a single illustrative run. 
We instead qualitatively highlight distinct search dynamics across the four acquisition strategies, supporting our discussion in Section~\ref{methods} regarding the different acquisition mechanisms of \textit{q}HAX-$\alpha$ compared with expected hypervolume improvement-based and scalarization-based strategies. Specifically, although all four methods share an identical surrogate posterior at Epoch~0, the candidate batches selected by \textit{q}HAX-UCB and \textit{q}HAX-EI diverge immediately from those of \textit{q}LNEHVI and \textit{q}LNParEGO across Epochs~0--5, visually confirming that decoupling per-objective exploration in acquisition space from posterior-mean hypervolume ranking induces search dynamics that are fundamentally distinct from both noisy hypervolume integration and randomized Chebyshev scalarization. The visualization also illustrates the differing behaviors of the \textit{q}HAX-$\alpha$ variants, driven by the inherent exploration-exploitation tendencies of their underlying single-objective strategies. At Epochs~0 and~1, the \textit{q}HAX-UCB variant identifies and samples the true Pareto set $\mathcal{X}^\star$ in the upper-left corner while simultaneously placing an exploratory candidate along the upper boundary. The UCB acquisition component drives systematic uncertainty reduction across the unvisited corners of the domain throughout subsequent epochs. By contrast, \textit{q}HAX-EI concentrates its early batch evaluations tightly along $\mathcal{X}^\star$, echoing the behavior of classical Expected Improvement, which decays rapidly away from the current best observations and can promote more localized exploitation. Finally, across a single batch, the $Q$ candidates also show diversity in the normalized decision space, rather than clustering in the same region.  Overall, these results demonstrate that \textit{q}HAX-$\alpha$ can instantiate multi-objective acquisition policies with genuinely distinct search dynamics directly from classical single-objective acquisitions, without requiring bespoke multi-objective derivations or specialized optimization machinery.

\subsection{MO-Hedge Ablations and Analysis}
\label{app:hedgeablations}

\begin{figure}[!t]
\includegraphics[width=0.985\linewidth]{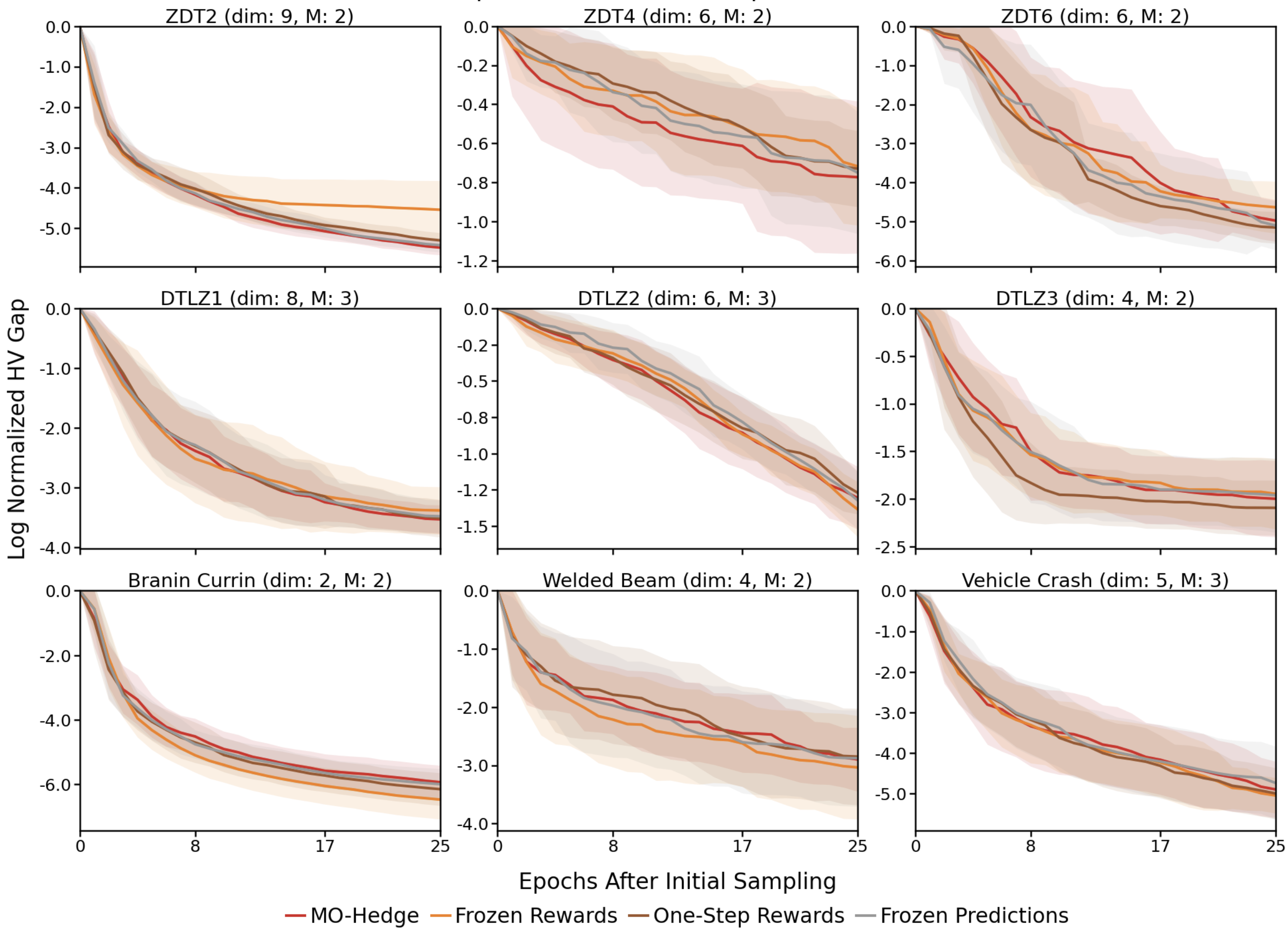}
\centering
\vspace{-2mm}
\caption{Log-normalized hypervolume gap for MO-Hedge variants using batch setting (\(Q=3\)). Lower values are better.
Mean (solid line) \(\pm 1\) standard deviation (shaded) over 20 runs.}
\label{fig:ablatehv}
\vspace{-3mm}
\end{figure}

\paragraph{Reward Formulation Ablations:}
As discussed in Section~\ref{methods}, MO-Hedge defines the historical reward for each constituent acquisition strategy $\alpha_k$ at iteration $t$ by evaluating the hypervolume of its cumulative candidate history $\mathcal{H}_t^k$ under the current standardized GP posterior mean $\boldsymbol{\mu}_t$:
\[
g_t^k = \mathrm{HV}\!\big(\{\boldsymbol{\mu}_t(\mathbf{x}) : \mathbf{x} \in \mathcal{H}_t^k\};\, \mathbf{r}^{\mathrm{hedge}}\big).
\]
Because $\boldsymbol{\mu}_t$ is updated as new observations are collected, recalibrating $\mathcal{H}_t^k$ at every iteration ensures that historical nominees are judged using the most recent available surrogate, rather than locking in potentially inaccurate predictions from early iterations when data were sparse. To evaluate the empirical impact of this design choice, Figure~\ref{fig:ablatehv} compares the default posterior-recalibrated MO-Hedge reward (instantiated with the MO-Hedge~1 portfolio) against three alternative reward formulations across all nine synthetic benchmarks ($Q=3$, 20 paired runs):

\begin{enumerate}
    \item \textit{Frozen Rewards:} At each iteration $t$, computes the posterior-mean hypervolume of only the newly proposed batch $\mathcal{C}_{t-1}^k = \mathcal{H}_t^k \setminus \mathcal{H}_{t-1}^k$ under $\boldsymbol{\mu}_t$ and adds it incrementally to the previous cumulative gain, $g_t^k = g_{t-1}^k + \mathrm{HV}(\boldsymbol{\mu}_t(\mathcal{C}_{t-1}^k);\, \mathbf{r}^{\mathrm{hedge}})$. As noted in the Theoretical Remark in Section~\ref{methods}, this stationary accumulation matches the classical GP-Hedge structure~\cite{brochu2011portfolio} required for standard $\mathcal{O}(\sqrt{T})$ expert-selection regret bounds.
    \item \textit{One-Step Rewards:} Computes $g_t^k = \mathrm{HV}(\boldsymbol{\mu}_t(\mathcal{C}_{t-1}^k);\, \mathbf{r}^{\mathrm{hedge}})$ using only the candidates proposed at the immediately preceding iteration, discarding all earlier candidate history.
    \item \textit{Frozen Predictions:} Evaluates the hypervolume over the full cumulative history $\mathcal{H}_t^k$, but freezes the objective prediction of each historical nominee $\mathbf{x} \in \mathcal{C}_{\tau-1}^k$ at the posterior mean $\boldsymbol{\mu}_\tau(\mathbf{x})$ computed at the iteration $\tau \le t$ when it was first evaluated, rescaled to the current standardized objective space at iteration $t$, without re-querying the updated GP posterior $\boldsymbol{\mu}_t$.
\end{enumerate}

As shown in Figure~\ref{fig:ablatehv}, all four reward formulations achieve strong and closely matched performance across the benchmark suite, demonstrating that hypervolume-based portfolio selection is robust to the specific credit-assignment variant. Nevertheless, the default posterior-recalibrated \texttt{MO-Hedge} formulation achieves the strongest aggregate performance across the nine benchmarks, with a clear advantage over \texttt{Frozen Rewards} on ZDT2 and ZDT4 and over \texttt{Frozen Predictions} on Vehicle Crash. Under \texttt{Frozen Rewards}, large gains accumulated during the first few iterations, when the surrogate is poorly calibrated, persist indefinitely in the cumulative sum and can delay adaptation once the surrogate improves, an issue that similarly motivated reward discounting in single-objective portfolio allocation~\cite{brochu2011portfolio}. Conversely, \texttt{One-Step Rewards} removes all historical bias and performs competitively on several problems, but discards cumulative Pareto-front coverage information across iterations. Posterior recalibration over $\mathcal{H}_t^k$ balances these trade-offs by retaining full historical coverage while continuously correcting early surrogate errors as new data arrive.

\begin{figure}[!b]
\includegraphics[width=0.985\linewidth]{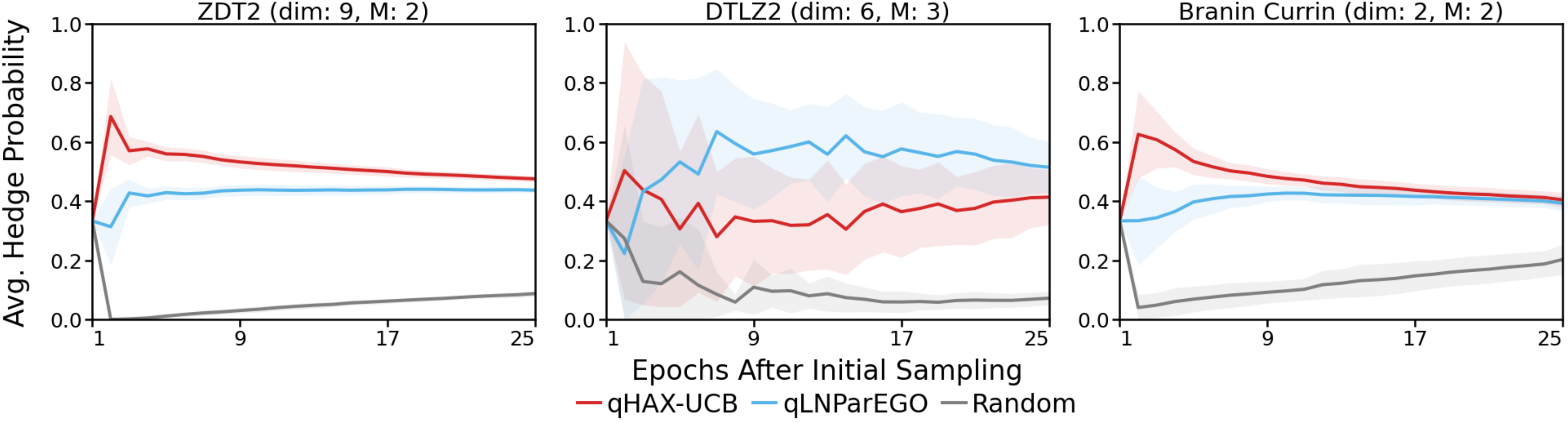}
\centering
\vspace{-2mm}
\caption{Average MO-Hedge selection probabilities for the acquisition strategies in the portfolio using batch setting (\(Q=3\)). Mean selection probability (solid line) \(\pm 1\) standard deviation (shaded) over 20 runs.}
\label{fig:hedgeprob}
\vspace{-3mm}
\end{figure}

\paragraph{Evolution of Portfolio Selection Probabilities:}
To directly illustrate how MO-Hedge adapts its acquisition preferences online, Figure~\ref{fig:hedgeprob} plots the mean selection probabilities $p_t^k$ ($\pm 1$ standard deviation across 20 runs) for the three constituent strategies in MO-Hedge~1 (\textit{q}HAX-UCB, \textit{q}LNParEGO, and uniform \texttt{Random} sampling) over 25 batch optimization epochs ($Q=3$) on three representative benchmarks: ZDT2, DTLZ2, and Branin--Currin. At Epoch~1, prior to observing any historical nominees ($\mathcal{H}_0^k = \emptyset$), all three strategies start from the uniform initialization $p_1^k = 1/3$. By Epoch~2, as soon as the first proposed batches are evaluated under the surrogate posterior, the unnormalized exponentiated-weights update sharply downweights \texttt{Random} sampling toward zero across all three benchmarks, confirming that the hypervolume reward immediately filters out uninformative policies.

The subsequent evolution of $p_t^k$ reflects the problem-dependent strengths of the constituent acquisitions observed in Figure~\ref{fig:q3hv}. On ZDT2 and Branin--Currin, where \textit{q}HAX-UCB achieves faster hypervolume convergence than \textit{q}LNParEGO, MO-Hedge assigns the highest selection probability to \textit{q}HAX-UCB throughout optimization. In contrast, on DTLZ2, where scalarization via \textit{q}LNParEGO outperforms \textit{q}HAX-UCB in Figure~\ref{fig:q3hv}, MO-Hedge rapidly reverses its preference after the initial epochs and concentrates nearly $60\%$ of its probability mass on \textit{q}LNParEGO. Finally, the selection probabilities gradually soften in later epochs, particularly on low-dimensional problems such as Branin--Currin ($d=2$). This behavior arises from two mechanisms: (i) the learning rate $\eta_t = \sqrt{8 \log K / n_t}$ decays as the dataset size $n_t$ grows, naturally relaxing the softmax distribution as optimization progresses, a behavior that can be tuned via alternative learning-rate schedules if desired, particularly since the HV-based formulation prevents unbounded reward aggregation with epochs; and (ii) as the cumulative history $\mathcal{H}_t^{\mathrm{Rand}}$ grows to $75$ uniform samples in a low-dimensional domain, a few random draws inevitably land near the Pareto set and increase the union hypervolume $g_t^{\mathrm{Rand}}$, which is particularly evident on Branin--Currin where $75$ uniform samples across a $d=2$ domain give a reasonable chance of Pareto-set detection. Overall, the evolution of the probability mass across iterations confirms that MO-Hedge rapidly suppresses uninformative strategies while dynamically reallocating evaluations toward the acquisition policy best suited to the underlying problem geometry, providing a principled and self-adapting methodology against acquisition mismatch in multi-objective Bayesian optimization.


\end{document}